\documentclass[journal]{IEEEtran}
\ifCLASSINFOpdf
\else
\fi
\usepackage{cite}
\usepackage{amsmath,amssymb,amsfonts}
\usepackage{algorithmic}
\usepackage{graphicx}
\usepackage{multirow}
\usepackage{textcomp}
\usepackage{hyperref}
\usepackage{filecontents}
\usepackage{lastpage}%
\usepackage{supertabular}%
\usepackage{tabu}%
\usepackage{multirow}%
\usepackage{diagbox}
\usepackage{longtable}%
\usepackage{color, colortbl}
\definecolor{Gray}{gray}{0.9}
\usepackage[algo2e,linesnumbered,ruled]{algorithm2e} 
\usepackage[numbers, sort&compress]{natbib}
\usepackage[flushleft]{threeparttable}
\usepackage[dvipsnames]{xcolor}

\begin{document}
\title{A Survey on Evolutionary Computation for Computer Vision and Image Analysis: Past, Present, and Future Trends}

\author{Ying Bi,~\IEEEmembership{Member,~IEEE}, Bing Xue,~\IEEEmembership{Senior Member,~IEEE}, Pablo Mesejo, Stefano Cagnoni,~\IEEEmembership{Senior Member,~IEEE}, Mengjie Zhang,~\IEEEmembership{Fellow,~IEEE}, 
\thanks{The author Ying Bi is with School of Electrical and Information Engineering, Zhengzhou University, Zhengzhou, China, and also with School of Engineering and Computer Science, Victoria University of Wellington, Wellington, New Zealand (e-mail: ying.bi@ecs.vuw.ac.nz).}% <-this % stops a space
\thanks{The authors Bing Xue and Mengjie Zhang are with School of Engineering and Computer Science, Victoria University of Wellington, Wellington, New Zealand (e-mail: bing.xue@ecs.vuw.ac.nz; mengjie.zhang@ecs.vuw.ac.nz).}% <-this % stops a space
\thanks{The author Pablo Mesejo is with Department of Computer Science and Artificial Intelligence (DECSAI), University of Granada, and also with the Andalusian Research Institute in Artificial Intelligence (DaSCI) and Panacea Cooperative Research S.Coop., Granada, Spain (e-mail: pmesejo@ugr.es).}
\thanks{The author Stefano Cagnoni is with Department of Engineering and Architecture, University of Parma, Italy (e-mail: stefano.cagnoni@unipr.it).}
\thanks{Color versions of one or more of the figures in this paper are available
online at http://ieeexplore.ieee.org.}}
\markboth{IEEE transactions on evolutionary computation,~Vol.~XX, No.~X, Month~year}
{Bi \MakeLowercase{\textit{et al.}}: paper title}
\maketitle

\begin{abstract} 
Computer vision (CV) is a big and important field in artificial intelligence covering a wide range of applications. Image analysis is a major task in CV aiming to extract, analyse and understand the visual content of images. However, image-related tasks are very challenging due to many factors, e.g., high variations across images, high dimensionality, domain expertise requirement, and image distortions. Evolutionary computation (EC) approaches have been widely used for image analysis with significant achievement. However, there is no comprehensive survey of existing EC approaches to image analysis. To fill this gap, this paper provides a comprehensive survey covering all essential EC approaches to important image analysis tasks including edge detection, image segmentation, image feature analysis, image classification, object detection, and others. This survey aims to provide a better understanding of evolutionary computer vision (ECV) by discussing the contributions of different approaches and exploring how and why EC is used for CV and image analysis. The applications, challenges, issues, and trends associated to this research field are also discussed and summarised to provide further guidelines and opportunities for future research. 

%The challenges and future directions seek more junior researchers to join this field to tackle with. 
\end{abstract}
\IEEEpeerreviewmaketitle

\section{Introduction}
Computer vision (CV) is an important and well-established research field that studies the use of machines and/or computers to mimic human vision systems and acquire, extract, understand, and analyse information from images and videos \cite{forsyth2011computer}. CV has many fundamental applications in various fields, such as security, remote sensing, engineering, biology, and medicine \cite{szeliski2010computer}. Image analysis is a major task in CV that aims to extract and analyse meaningful information from images. A wide range of CV and image analysis tasks, such as image segmentation, image feature extraction, image classification, face recognition, object detection, object tracking, have been studied for decades \cite{bhuyan2019computer}. However, these tasks are still very challenging due to a number of reasons, such as data complexity, computational cost, low interpretability, lack of sufficient labelled data, high data dimensionality, and high variation across images. Therefore, CV is still a very rapidly developing area, where many new research and techniques are being proposed to effectively solve different tasks. 

Artificial intelligence (AI) covers a variety of techniques that simulate human intelligence to solve different tasks \cite{russell2002artificial}. Evolutionary computation (EC) includes a family of population-based techniques under the big umbrella of AI \cite{bi2021gpimage}. Typical EC techniques are evolutionary algorithms, swarm intelligence and others \cite{xue2015survey}. EC techniques have successfully solved many tasks related to engineering, finance, medicine, biology, management, business, manufacturing, and remote sensing \cite{zhan2022survey}. The tasks solved by EC range from optimisation to learning, including function/parameter optimisation, classification, regression, and clustering. 

Their population-based characteristics make EC techniques applicable and effective for many CV and image analysis tasks without requiring extensive domain knowledge while providing optimal or at least human-competitive solutions \cite{cagnoni2016evolutionary, olague2016evolutionary}. EC techniques use one or more populations to search for optimal solutions through a number of generations guided by one or more fitness/objective functions. The population-based search provides EC techniques with powerful search ability and problem-dependent fitness functions to guide the evolutionary process towards promising solutions. Many image analysis tasks, such as image feature selection and extraction, typically have a large search space within which manually designed solutions and exact methods may not be effective. EC techniques can provide high-quality and sometimes human-interpretable solutions to these tasks. Furthermore, EC techniques can well handle problems that have multiple conflicting objectives by producing a set of non-dominated solutions. These problems, which cannot be easily solved using exact methods, are known as multi-objective optimisation problems.

\begin{figure}
	\centering
	\includegraphics[width=\linewidth]{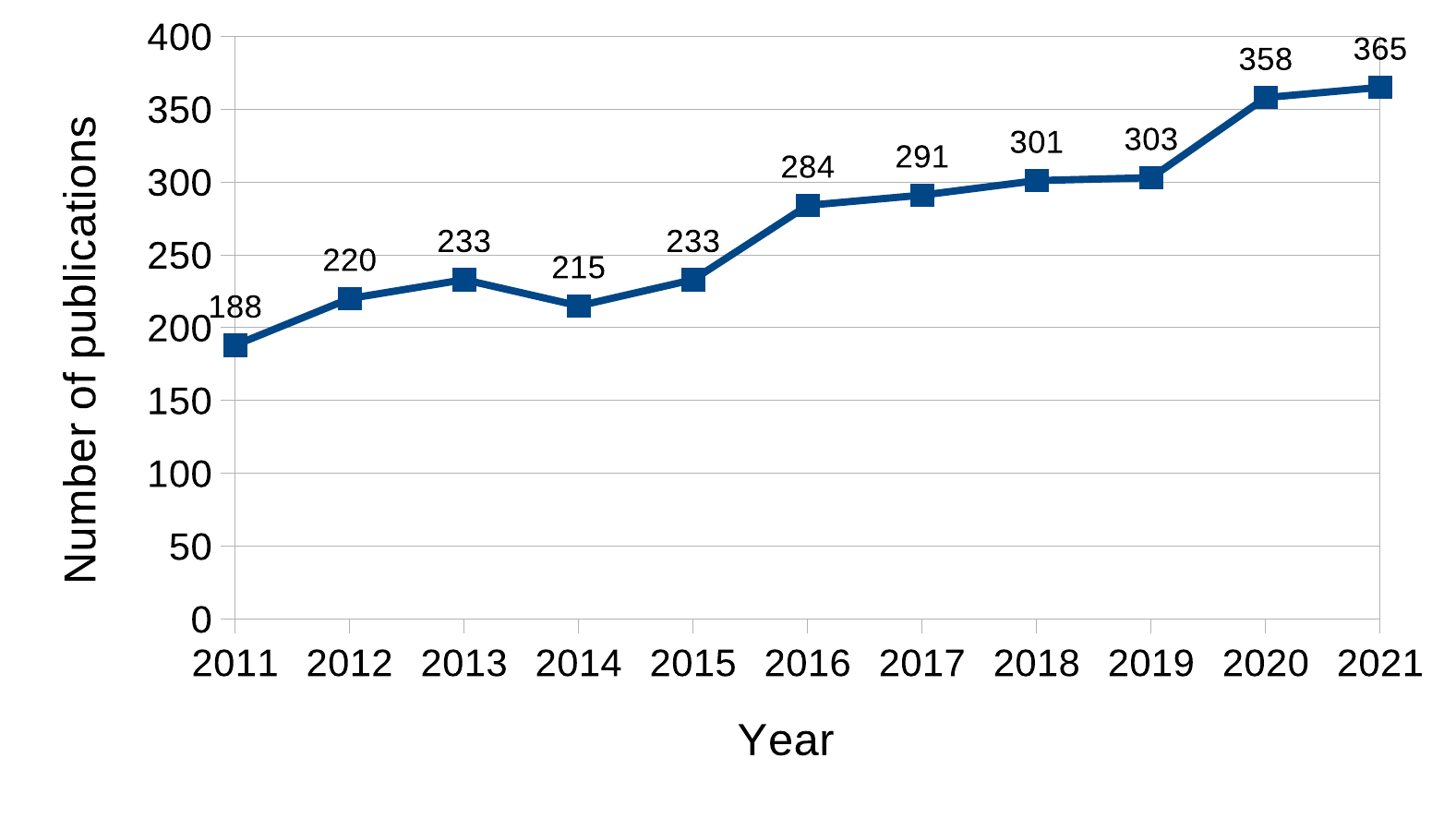}
	\caption{The number of publications related to ECV from 2011 to 2021. The data is collected from Scopus with the keywords of ``evolutionary" AND ``computer vision" OR ``image analysis". The total number of publications from 2011 to 2021 is 4955.}
	\label{fig:no_paper_peryear}
	% \vspace{-4mm}
\end{figure}

Since the 1970s, EC techniques have been successfully applied to CV and image analysis \cite{cagnoni2016evolutionary}. This field has also been termed evolutionary computer vision (ECV) in the last decade \cite{olague2016evolutionary}. It is worth mentioning that Gustavo Olague published the first authored book on ECV in 2016 \cite{olague2016evolutionary}. Recent research on ECV has been published in fully refereed journals and annually held conferences related to EC, soft computing, CV, image processing, and pattern recognition. This is a fast-developing research field in which the number of publications has been gradually increasing in the recent decade, as shown in Fig. \ref{fig:no_paper_peryear}. There have been several ``specialised" surveys related to this topic, but they focus on particular tasks/aspects. \citeauthor{nakane2020application} \cite{nakane2020application} review typical works on genetic algorithms (GAs), differential evolution (DE), particle swarm optimisation (PSO), and ant colony optimisation (ACO) for CV tasks, including image matching, visual tracking, face recognition. However, that survey mainly focuses on applications and ignores other important EC methods including genetic programming (GP) and evolutionary multi-objective optimisation (EMO). %\citeauthor{khan2021recent} \cite{khan2021recent} review the applications of GP for image processing, ignoring other EC methods. 
\citeauthor{cagnoni2016evolutionary} \cite{cagnoni2016evolutionary} discuss questions, issues, challenges and future directions of EC for CV and image processing. However, there has been no comprehensive survey covering all recent ECV techniques. Therefore, this paper fills this gap by providing a comprehensive survey on this topic. 

%This shows the importance and popularity of the field of EC for image analysis. %Figure \ref{} also shows the trend of the number of publications 

This survey reviews the main trends and algorithms based on EC techniques for CV and image analysis, dealing with edge detection, image segmentation, image feature analysis, image classification, object detection, and other tasks, according to a task-based taxonomy. This field is developing very fast and many works on this topic have been published recently. Figure \ref{fig:no_paper_peryear} summarises the number of total publications from 2011 to 2021 on EC for CV and image analysis extracted from Scopus \footnote{\url{https://www.scopus.com}}. 
% and Fig. \ref{fig:no_paper_peryear_survey} shows the categories of the publications in the last five years. %Unsurprisingly, the largest number of publications regards EC for image segmentation and image classification. Figure \ref{fig:no_paper_peryear} also shows the number of reviewed publications per year. 
This survey paper reviews over 150 publications from databases such as IEEE Xplore, ACM Digital Library, Scopus, Springer, and Google Scholar. The selected publications meet at least one of the following criteria: 1) they are published in international journals and major conferences with high reputation and relevance in the last ten years (from 2012 to 2022); 2) they have been frequently cited; 3) they are from well-recognised researchers or research groups in this field.  Specifically, the survey paper includes over 140 publications from 2012 to 2022 and about 100 publications from 2017 to 2022. The survey provides a comprehensive overview of these works, summarises the application areas, challenges and issues, and highlights future research directions. 

% \begin{figure}
% 	\centering
% 	\includegraphics[width=\linewidth]{figure/Scopus-Analyze-Year2}
% 	\caption{The number of publications related to ECV from 1970 to 2021. The data is collected from Scopus with the keywords of ``evolutionary" AND ``computer vision" OR ``image analysis". The total number of publications from 1970 to 2021 is 2991.}
% 	\label{fig:no_paper_peryear}
% 	% \vspace{-4mm}
% \end{figure}

\section{Taxonomy and Scope}

\begin{figure}
	% \vspace{-4mm}
	\centering
	\includegraphics[width=\linewidth]{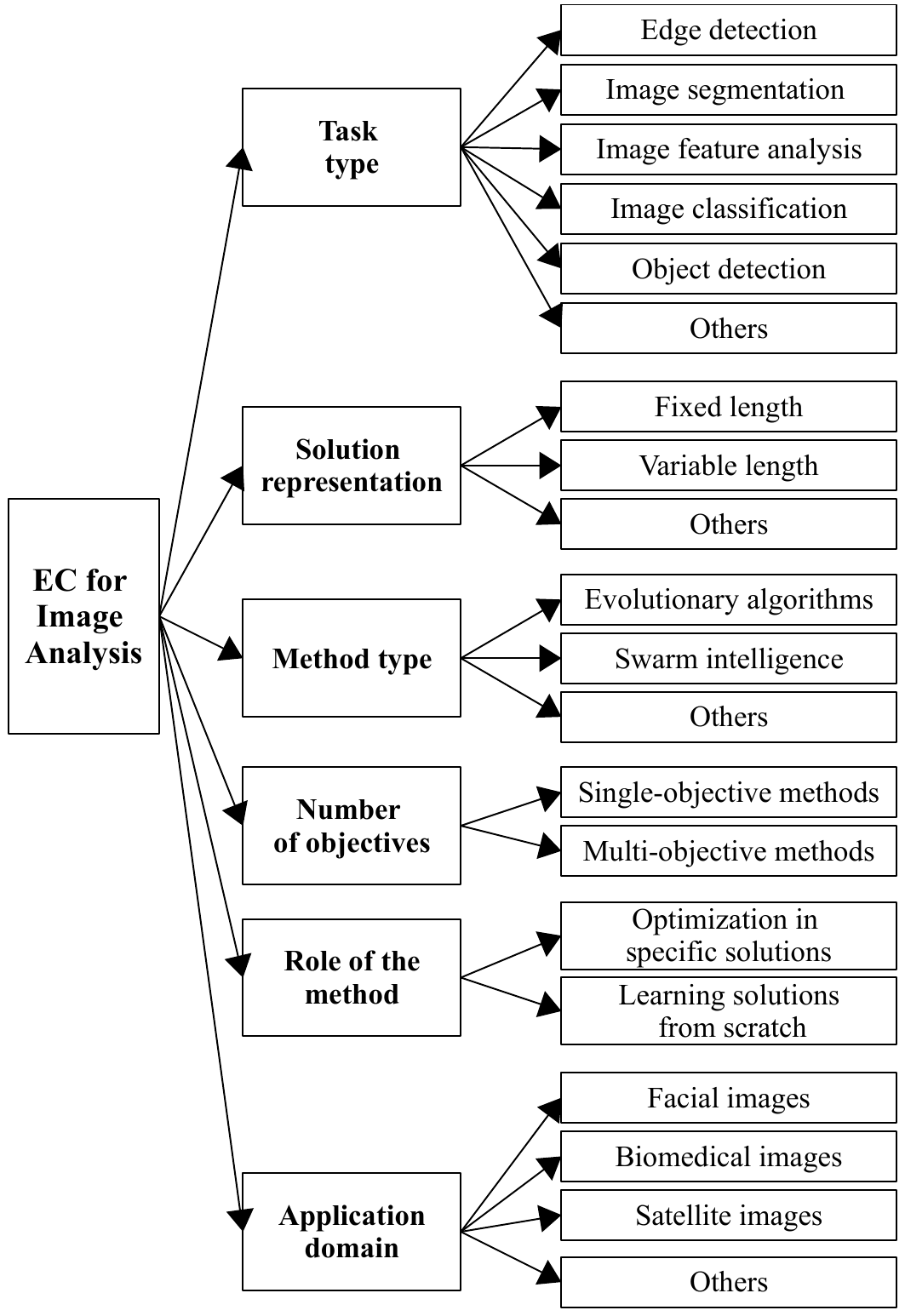}
	% \vspace{-6mm}
	\caption{Categories of existing EC methods for image analysis.}
	\label{fig:eciaflowchat}
	% \vspace{-2mm}
\end{figure}

Existing EC approaches to CV and image analysis, as shown in Fig. \ref{fig:eciaflowchat}, can be categorised into the following groups based on different criteria, i.e.,

\begin{itemize}
	\item edge detection, image segmentation, image classification, object detection, etc, according to \emph{the task type};
	\item flexible variable-length representation and fixed-length string or vector-based numeric representation, according to \emph{the solution representation};
	\item GAs, GP, PSO, ACO, DE, EMO, etc, according to \emph{the method};
	\item single-objective optimisation methods and multi-objective optimisation methods, according to \emph{the number of objectives};
	\item optimisation of specific solutions/models and learning/constructing models from scratch, according to \emph{the role of the EC method};
	\item facial image analysis, biomedical image analysis, remote sensing/satellite image analysis, etc, according to \emph{the application domain}.
\end{itemize}

%\begin{figure}[htbp]
%	\centering
%	\includegraphics[width=\linewidth]{figure/structure_survey2}
%	\caption{Structure of the survey paper.}
%	\label{fig:structuresurvey}
%		\vspace{-4mm}
%\end{figure}

This paper follows a task-based taxonomy for a better understanding of how EC is used to solve different tasks, including edge detection, image segmentation, image feature analysis, image classification, and object detection, which are representative CV and image analysis tasks. A summary of the reviewed works on EC for these tasks is listed in Table \ref{table:ecSummary}.

\begin{table}
		% \vspace{-2mm}
	\footnotesize
	\setlength{\tabcolsep}{0.4em} % for the horizontal padding
	\caption{Summary of EC approaches to Image Analysis}
	% \vspace{-4mm}
	\begin{center}
		\begin{tabular}{|p{0.19\linewidth}|p{0.48\linewidth}|p{0.24\linewidth}|}
			\hline 
			Task&EC-based approach& References\\
			\hline	
			\multirow{2}{\hsize}{Edge detection}&Optimise existing methods&\cite{sengupta2019improved, lu2008edge, setayesh2013novel,  bacsturk2009efficient, zheng2019differential}\\ 
			&Construct solutions from scratch&\cite{fu2013low, fu2018fast, fu2019bayesian, nezamabadi2006edge, tian2008ant, kumar2020edge, pajouhi2017image, fu2016genetic}\\ \hline
			\multirow{5}{\hsize}{Image segmentation}&Threshold-based methods& \cite{rundo2019novel, li2015dynamic, zhao2016multilevel, suresh2017multilevel, li2017partitioned, mirghasemi2019domain, zhao2021chaotic, zhao2021ant, sarkar2013multilevel, ali2014multi, ayala2015image, bhandari2020novel, muangkote2017rr, tarkhaneh2019adaptive, abd2019many}\\
			&Region-based methods&\cite{xie2013automatic, gong2017differential, lee2012saliency, rogai2016metaheuristics, liu2021multiobjective}\\
			&Clustering-based methods& \cite{khan2015genetic, singh2019sunflower, khan2015modified, das2009automatic, das2010kernel, zhao2018noise, zhao2020semisupervised}\\
			&Classification-based methods&\cite{poli1996genetic, song2008texture, liang2014image, liang2015supervised, liang2017genetic, liang2019figure, liang2020evolving, zhou2019evolutionary, hassanzadeh2020evolutionary, hassanzadeh2020evou, wei2021genetic, lima2021automatic}\\
 			&Others&\cite{ayala2006circle, amelio2014evolutionary, mesejo2012automatic}\\
			\hline
			\multirow{2}{\hsize}{Image feature analysis}&Feature selection&\cite{lin2014study, alirezazadeh2015genetic, kirar2020combination, hemanth2019modified, ain2017genetic, naeini2018particle, tan2018intelligent, tan2019intelligent, kavuran2021sem, rashno2017effective, sweetlin2017feature, devarajan2020metaheuristic, ghosh2013self, mlakar2017multi, liang2017image, thangavel2014soft, ghamisi2014feature, mistry2016micro}\\
			&Feature extraction and learning&\cite{bi2021gpimage, albukhanajer2014evolutionary, gong2015multiobjective, albukhanajer2017classifier, krawiec2005visual, krawiec2007visual, perez2009evolutionary, perez2013genetic, al2016automatically, al2017keypoints, al2020automatically, rodriguez2018structurally, rodriguez2019evolving, rodriguez2020cooperative, liu2015learning, liu2015sequential, bi2018gaussian, bi2018genetic, bi2020effective, bi2019tevc, bi2020automatically, bi2021multi, bi2021divide, bi2021instance}\\
			\hline
			\multirow{2}{\hsize}{Image classification}&Evolving NN-based Methods& \cite{sun2019evolving, chen2020evolving, o2021evolutionary, real2019regularized, zhang2022evolutionary, zhang2021adaptive, li2021automatic, suganuma2017genetic, li2019evolving, gong2020evolving, zhang2020efficient, lu2020nsganetv2, sun2019surrogate, wang2021surrogate, lu2020multiobjective, lu2021neural, wen2021two, zhu2019multi, zhu2021real, wang2019evolving}\\
			&Evolving non-NN-based Methods&\cite{choi2012genetic, ryan2015image, ghazouani2021genetic, smart2003classification, zhang2006using, al2012two, lensen2016genetic, burks2018genetic, bi2018automatic, fan2020region, fan2022genetic, bi2020genetic, bi2020evolving, plichoski2021face, shao2013feature, agapitos2015deep, ibarra2022brain, olague2017brain, al2016binary, bi2021dual, bi2022using, iqbal2017cross, bi2021learning}\\
			\hline
			\multirow{2}{\hsize}{Object detection}&Optimising object detection systems&\cite{ganesh2014entropy, abdel2014efficient, ugolotti2013particle, mussi2009gpu, mussi2010gpu, singh2014novel, iqbal2016learning, afzali2017supervised, moghaddam2021automatic}\\
			&Automatically evolving detectors& \cite{howard1999target, howard2006pragmatic, bhanu2004object, zhang2003domain, zhang2007improving, liddle2010multi, li2021re, olague2022automated}\\
			\hline
			\multirow{5}{*}{Others}	&Interest point detection&\cite{trujillo2006synthesis, trujillo2008automated, olague2011evolutionary, olague2012interest}\\
			&Image registration&\cite{santamaria2011comparative, valsecchi2013evolutionary, santamaria2012self, gomez20183d} \\
			&Remote sensing image classification&\cite{mai2021hybrid, wang2017remote, liu2020multiobjective, rs14051275}\\
			&Object tracking&\cite{kang2018hybrid, song2013understanding, yan2021lighttrack}\\		
			%&\colorbox{Apricot}{aa}&\colorbox{Green}{aa}&\colorbox{Lavender}{aa}&\colorbox{Cyan}{aa}&\colorbox{GreenYellow}{aa}&\colorbox{Apricot}{aa}&\colorbox{Apricot}{aa}\\
			\hline		
		\end{tabular}	
		\label{table:ecSummary}
	\end{center}
	% \vspace{-6mm}
\end{table}

EC techniques have also been applied to other CV and image analysis tasks, e.g., video games \cite{stanley2005real}, image retrieval \cite{torres2009genetic}, scene change detection \cite{bianco2017combination}, or image reconstruction \cite{jiao2016novel}. Due to the page limit, those works cannot be covered in this survey. Also notice that this paper focuses on CV and image analysis, while image processing tasks such as image enhancement, image restoration and image compression are beyond the scope of this paper. As well, this review does not include all the CV tasks, since the applications of EC methods to some of them are very limited.

\section{Edge Detection} \label{edge_detection}

\begin{figure}[htbp]
	\centering
	% \vspace{-6mm}
	\includegraphics[width=\linewidth]{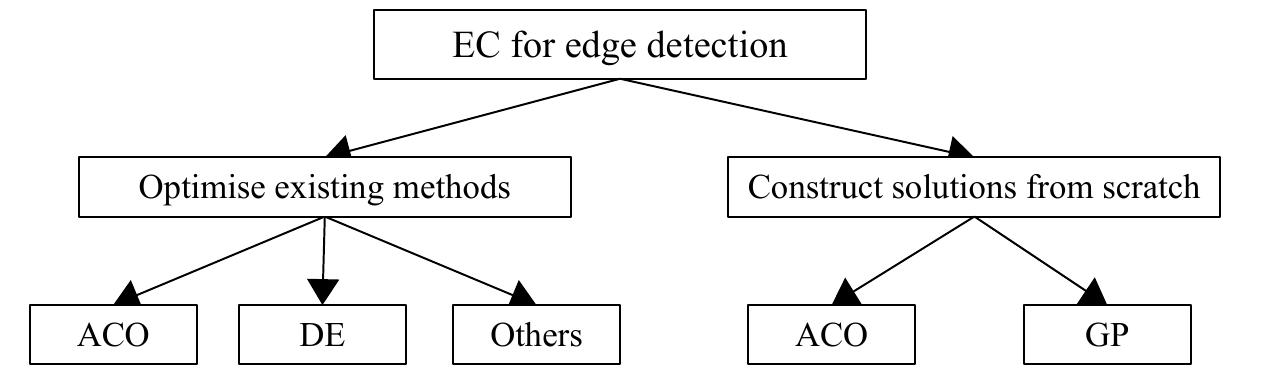}
	% \vspace{-6mm}
	\caption{EC techniques used for edge detection.}
	\label{fig:edgedetection}
	% \vspace{-2mm}
\end{figure}

Edge detection is the task of finding or detecting discontinuities of pixel values in the image. A simple example solution is to compare the value of the current pixel with the values of its neighbouring pixels from different directions to approximate the pixel value gradient. If it changes significantly, the current pixel will possibly be an edge pixel. Edge detection is an important task in CV and image processing to obtain low-level image features. Edge detection is also important for many image analysis tasks, such as object detection and image segmentation. However, due to the complex background and noise in images, edge detection is challenging. EC techniques have been widely applied for edge detection, as shown in Fig. \ref{fig:edgedetection}. 

%(edge detection images: Camera, House, Lena, Pepper, Kid, Fruit, Home, )

%\begin{table}[htbp]
%	%	\vspace{-4mm}
%	%	\footnotesize
%	%\renewcommand{\arraystretch}{1.13}
%	\caption{Summary of EC approaches to edge detection}
%	%	\vspace{-4mm
%	\begin{center}
%		\begin{tabular}{p{0.58 \linewidth}|p{0.42 \linewidth}}
%			\hline 
%			Categories&References \\ \hline 
%			Construct solutions from scratch using ACO& \cite{nezamabadi2006edge, tian2008ant, kumar2020edge, pajouhi2017image}
%			\\ \hline 
%			Automatically learn edge detectors using GP&\cite{fu2013low, fu2018fast, fu2016genetic, fu2019bayesian, fu2014feature}
%			\\ \hline
%			Optimise results of existing edge detectors& \cite{lu2008edge, wong2008improved, sengupta2019improved}
%			\\ \hline 
%			Optimise parameters of existing algorithms& \cite{bacsturk2009efficient} \cite{verma2016optimal} \cite{zheng2019differential} 
%			\\		\hline	
%		\end{tabular}	
%		\label{table:edge_detection}
%	\end{center}
%	%	\vspace{-6mm}
%\end{table}

ACO is the most commonly used method for edge detection using a graph-based representation. An image can be represented by a graph where a node represents a pixel. Ants can move from one node to another node and mark the node by increasing the corresponding cell in a ``pheromone" matrix. Edges can be detected by setting a threshold on the pheromone deposited by ants. In the ACO methods for edge detection, the movement of ants from one node to another and the pheromone matrix update are the main operations. Early works on ACO for edge detection can be found in \cite{nezamabadi2006edge, tian2008ant}, that present different strategies of moving ants. 
%\citeauthor{nezamabadi2006edge} \cite{nezamabadi2006edge} propose an ACO method with two node transition rules are used, i.e., moving an ant to one of its eight neighbours or dispatching it to a random node. A global threshold is used to discard irrelevant edges and a thinning method is used to further improve the edge detection results. \citeauthor{tian2008ant} \cite{tian2008ant} propose an ACO method that moves ants to one of its four or eight neighbour nodes according to the local variations of the image's intensity values. In this method, an adaptive threshold that considered two classes of edge and non-edge pixels is used to obtain the edge detection results. This method is tested on four images without using any post-processing method.
 \citeauthor{kumar2020edge} \cite{kumar2020edge} apply an advanced version of a bilateral filter to suppress image artifacts and use ACO to detect edges. The results show that guided image filtering can improve the ACO-based edge detection results by detecting some non-prominent edges.
%for image edge detection. This method used a population of ants to search for the solution and established a pheromone information matrix, which contains the information of edges and non-edges in the image. This method also used the information of neighbouring pixels value to guide the solution construction and the movement of ants. The superiority of this method has shown on several images over other edge detection methods, including ACO-based methods. \citeauthor{kumar2020edge} \cite{kumar2020edge} propose an ACO method to detect edges from images. Similar to \cite{tian2008ant, nezamabadi2006edge}, the ants ar placed in the image randomly and search for edge or non-edge pixels by updating the pheromone matrix. In this method, all the possible neighbourhood pixels are considered when moving ants from one node to another node. The results showed that ACO failed to detect some non-prominent edges, while this is improved by using guided image filtering before edge detection. 
\citeauthor{pajouhi2017image} \cite{pajouhi2017image} apply ACO to edge detection and implement the ACO algorithm using memristive networks. The edge detection process includes three steps, i.e., filtering, image enhancement, and edge detection. ACO is used to find the edge points in the image. Each ant can move to four neighbouring directions during the search process, updating the pheromone accordingly. %In addition, the implementation of ACO using memristive networks is presented to improve energy consumption. %This method achieves a 28\% improvement in the energy compared with the normal implementation. 

Instead of searching from scratch, ACO has also been used to improve the results obtained by traditional edge detection methods. \citeauthor{lu2008edge} \cite{lu2008edge} apply ACO to repair broken edges in the results obtained by two traditional edge detectors, i.e., Sobel and Canny. This method places ants where the edges break to extend edges based on four moving policies to reduce search redundancy. This method is effective and efficient in improving edge detection performance. %\citeauthor{wong2008improved} \cite{wong2008improved} propose an ACO method to find links between disjointed edges produced by Canny filtering. Ants are placed at endpoints to search for the best paths in the image generated by Canny to obtain the edge detection results. This method improves edge detection performance but requires careful parameter settings. 
\citeauthor{sengupta2019improved} \cite{sengupta2019improved} propose an ACO method for skin lesion edge detection by searching for edge contours in the results generated by Canny, Sobel or Prewitt. Combining ACO with Canny achieves the best results on three skin lesion images.

Besides ACO, GP can automatically construct models that classify pixels into the edge or non-edge groups or identify edge pixels from images. \citeauthor{fu2013low} \cite{fu2013low} develop a GP method with a new function set and a new program structure to evolve tree-based edge detectors on natural images. The evolved edge detectors mark a pixel as an edge or non-edge point based on a predefined threshold. The results show that the automatically constructed edge detectors are very effective and the leaf nodes (pixels) in the edge detectors are important. \citeauthor{fu2018fast} \cite{fu2018fast} develop an unsupervised GP method that do not require ground truth (training) images to automatically evolve edge detectors. This method uses a special tree structure, new function and terminal sets to evolve programs to classify pixels. It uses an energy function based on the average of the image gradients and the sum of image gradients as the GP fitness function. This method achieves significantly better results than a baseline GP method and the Canny edge detector. \citeauthor{fu2016genetic} \cite{fu2016genetic} apply GP to evolve the combinations of different Gaussian filters and their parameters for edge detection. This method uses a new fitness function based on the localisation accuracy. \citeauthor{fu2019bayesian} \cite{fu2019bayesian} develop a GP method to evolve Bayesian programs that construct a high-level feature for edge detection. %More related work on using GP to construct edge detectors from scratch can be found in \cite{fu2014feature}.

Other EC methods have been used to optimise the process of edge detection. 
\citeauthor{setayesh2013novel} \cite{setayesh2013novel} propose two PSO methods with different constraint-handling strategies to detect edges in images. Specifically, a particle is encoded by a number of  consecutive pixels representing the detected edges. This method achieves better performance than Canny and the robust rank order-based algorithm. In \cite{bacsturk2009efficient}, DE is shown to be an effective method to find optimal coefficients of the cloning template in a cellular neural network for edge detection. %Unlike ACO-based edge detection methods, this method needs training images to calculate the learning error, which is used as fitness in DE.
%\citeauthor{verma2016optimal} \cite{verma2016optimal} develop a fuzzy system with a histogram-based Gaussian membership function and a parametric fuzzy intensification operator for edge detection and employ bacterial foraging algorithm (BFA) to search for four parameters in this system by optimising a sharpness factor and fuzzy entropy. This method achieves more accurate edge detection results than many traditional edge detectors but requires longer computational time. 
\citeauthor{zheng2019differential} \cite{zheng2019differential} propose an improved DE method to generate better images as inputs of the generator of generative adversarial network (GAN) for edge detection. The fitness function of DE is the loss function calculated from the discriminator of the GAN. This method achieves state-of-the-art performance on two benchmark datasets. However, it requires a number of ground-truth images to train the models and is time-consuming.

To sum up, when used for edge detection, EC techniques can automatically evolve edge detectors, construct graphs for edge detection, optimise the results produced by traditional detectors, and optimise the parameters of existing algorithms. ACO with a graph-based representation and GP with a tree-based representation are more popular than other EC methods for edge detection. The unique representations of ACO and GP make them more suitable for designing or evolving solutions to edge detection from scratch. The ACO-based methods are typically applicable to a single image. In most GP-based methods and the methods optimising existing learning methods, training images are typically needed for learning the models/detectors. The learned models can be generalised to detect edges from different images. However, the potential of EC-based edge detection methods has not been fully investigated on complex real-world images. The combinations of EC with existing powerful learning methods and the use of EC to automatically construct edge detectors from scratch will be potential future directions for the complex edge detection tasks.

\section{Image Segmentation}
Image segmentation aims to divide an image into multiple non-overlapping regions each of which is homogeneous \cite{pal1993review}. Image segmentation is an important step in image analysis, often necessary to solve higher-level tasks such as image classification and object detection. Image segmentation is a difficult task that may involve complex or large search spaces, have high computational cost, and require rich domain knowledge. Typical image segmentation methods include region-based methods, clustering-based methods, threshold-based methods, graph-based methods, edge-based methods, and classification-based methods \cite{khan2014survey, bhandarkar1999image, liang2014image}. Existing methods can also be categorised according to the types of images, e.g., medical images, biology images, satellite images, and thermal images. 

\begin{figure*}[ht]
	\centering
	\includegraphics[width=\linewidth]{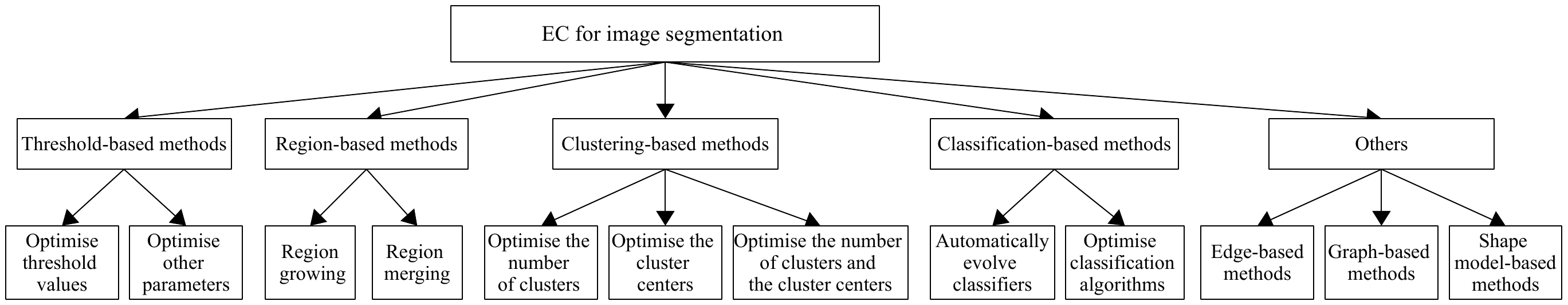}
	% \vspace{-6mm}
	\caption{Categories of existing EC-based approaches to image segmentation.}
	\label{fig:imagesegmentationOverall}
	% \vspace{-6mm}
\end{figure*}

EC has been widely applied to image segmentation because of the powerful global search ability and low requirement of domain knowledge. Early works on EC-based image segmentation can be dated back to the 1990s \cite{bhanu1994genetic, bhanu1995adaptive}. \citeauthor{liang2014image} \cite{liang2014image} review typical works on EC for image segmentation before 2014. In \cite{liang2014image}, EC-based image segmentation methods are broadly grouped into two types, i.e., using EC techniques to optimise existing segmentation algorithms and using EC techniques to automatically evolve models for image segmentation. \citeauthor{mesejo2016survey} \cite{mesejo2016survey} provide a survey on metaheuristics-based methods including EC methods for image segmentation with a special focus on the deformable model-based image segmentation. However, at least in the last five years, there is no systematic review on EC methods for image segmentation. This section will fill this gap by discussing recent works, categorised into five major classes, i.e., thresholding-based, region-based, clustering-based, classification-based, and others. The latter group of image segmentation methods include edge-detection-based methods \cite{ayala2006circle}, graph-based methods \cite{amelio2014evolutionary}, and model-based methods \cite{mesejo2012automatic}. The taxonomy is summarised in Fig. \ref{fig:imagesegmentationOverall}.

%feature-based --> a cluster
%region-based --> region growing, region merging
%graph-based --> a hybrid method using feature-based and spatial information
%based on evaluation function optimisation --> use an evaluation function incorporating the feature-based and/or spatial information.
%local search and global search method

\subsection{Threshold-based Image Segmentation Methods}
One of the simplest and popular approaches to image segmentation are threshold-based methods. The main idea is to calculate the histogram or other statistics of images and subdivide the images by comparing predefined threshold values with the histogram/statistical values and then grouping similar pixels. For example, a region (segment) may be formed by pixels with histogram values larger than a single predefined threshold. For complex images, multiple threshold values are often needed to achieve good segmentation performance. However, it is difficult to manually set appropriate threshold values for image segmentation due to the large (search) space and the unknown optimal number of threshold values. EC methods have been widely used to find optimal threshold values by optimising different objective functions. 

DE has been the most popular method in this category in the last decade. %\citeauthor{saraswat2013leukocyte} \cite{saraswat2013leukocyte} propose a two-phase leukocyte segmentation method based on DE, where the first DE method optimises the threshold values based on the intensity of each pixel and the second DE method optimises the threshold values based on the segmentation results of the first phase and four cell features, by maximising the inter-class variance. The proposed method outperforms traditional image segmentation algorithms. 
\citeauthor{sarkar2013multilevel} \cite{sarkar2013multilevel} propose a DE method to find multiple optimal threshold values based on the 2D histogram of image pixels by maximising the two-dimensional Tsallis entropy. This method achieves better convergence and performance than PSO and GA. \citeauthor{ali2014multi} \cite{ali2014multi} develop a synergetic DE method with opposition-based learning, a new mutation operator using the best base individual, and dynamic population updating. DE is used in two different threshold-based methods, i.e., to find optimal parameters for the Gaussian functions approximating the distributions of the pixel values by minimising the probability error, and to find optimal threshold values by maximising Kapur's entropy, respectively. \citeauthor{ayala2015image} \cite{ayala2015image} develop a beta-DE method using a beta probability distribution to tune the control parameters during the evolutionary process. The DE method is used to optimise histogram threshold values in the Otsu-based image segmentation method \cite{otsu1979threshold} by maximising the inter-class variance. In \cite{bhandari2020novel}, the beta DE method is applied to colour image segmentation by maximising the Kapur's entropy and the Tsallis entropy, respectively. %\citeauthor{mlakar2016hybrid} \cite{mlakar2016hybrid} propose a hybrid method based on self-adaptive DE and the reset strategy from the Cuckoo search to find the threshold values in Otsu that maximise the between-class variance.
\citeauthor{muangkote2017rr} \cite{muangkote2017rr} propose a DE method with a new mutation operator for Otsu on pseudo images, real images and satellite images by optimising the number of threshold values. \citeauthor{tarkhaneh2019adaptive} \cite{tarkhaneh2019adaptive} propose a new DE method with an adaptive scheme and new mutation strategies based on different distributions to search for multiple threshold values in Otsu for medical image segmentation. % \citeauthor{xu2019novel} \cite{xu2019novel} propose a hybrid method based on the dragonfly algorithm and DE to optimise the Otsu measure or cross-entropy measure for colour image segmentation.

PSO is also popular for finding the optimal threshold values for image segmentation. %\citeauthor{gao2013particle} \cite{gao2013particle} propose a PSO method with an intermediate disturbance strategy to improve its search ability and apply this method to find multiple optimal threshold values based on the Otsu measure and Kapur's entropy, respectively. This method gains better results than the other EC methods. 
\citeauthor{li2015dynamic} \cite{li2015dynamic} propose a quantum-behaved PSO method based on dynamic context cooperation for Otsu for medical image segmentation. This method achieves better results than several quantum-behaved PSO variants and traditional methods. \citeauthor{zhao2016multilevel} \cite{zhao2016multilevel} propose a modified PSO-based method with a fitness function based on the sum of the two-dimensional K-L divergences of different regions. PSO's performance is improved by using an adaptive factor in updating the position of particles to prevent premature convergence. This method achieves better performance than existing methods and PSO variants on colour images. \citeauthor{suresh2017multilevel} \cite{suresh2017multilevel} propose a PSO-based method adopting chaotic sequence mitigation and a cross-entropy or Tsallis entropy for satellite image segmentation. The performance of this method is investigated by evolving different numbers of threshold values. \citeauthor{li2017partitioned} \cite{li2017partitioned} propose a cooperative method and search space partitioning in quantum-behaved PSO to find optimal threshold values in Otsu for medical image segmentation. \citeauthor{mirghasemi2019domain} \cite{mirghasemi2019domain} propose a PSO method to find optimal threshold values for images transformed by the two-dimensional discrete wavelet method. %The Canny edge detector is used to create an edge map, which is aggregated with a threshold coefficient map. Finally, a clustering method is used to generate the segmented image. 
This method is effective for severely noisy image segmentation. 

Other EC methods have been developed to find optimal threshold values. %\citeauthor{hilali2020quantum} \cite{hilali2020quantum} applied a quantum GA (QGA), a standard GA, and a PSO-based method to optimise threshold values. Three different measures of entropy, i.e., R\'enyi entropy, Masi entropy and Shannon entropy, are used as objective functions in these methods for medical image segmentation. The PSO-based method performs the best, QGA being the runner-up. The results indicate the R\'enyi entropy as the most suitable objective function. 
\citeauthor{zhao2021chaotic} \cite{zhao2021chaotic} modify the continuous ACO method by using a random spare strategy and chaotic intensification strategy to improve its search ability for image segmentation based on a two-dimensional histogram and Kapur's entropy. \citeauthor{zhao2021ant} \cite{zhao2021ant} further improve the ACO-based method by proposing a horizontal and vertical crossover search mechanism to reduce randomness. Extensive experiments confirm the effectiveness of the new components in ACO. %\citeauthor{osuna2016image} \cite{osuna2016image} propose the allostatic optimisation (AO) method to optimise threshold values. This method uses Cauchy functions to approximate the distributions of the histograms of images and the Hellinger distance as a fitness function. This method performs better than Otsu. 
%\citeauthor{oliva2019image} \cite{oliva2019image} develop an electromagnetism-like optimisation algorithm that simulates attraction and repulsion mechanisms by minimising the cross-entropy. %\citeauthor{hinojosa2018thermal} \cite{hinojosa2018thermal} investigate different EC techniques such as ABC and DE to evolve threshold values for thermal image segmentation with the between-class variance and the Kapur's entropy as objective functions.

%EC methods have been proposed to optimise other parameters in the threshold-based image segmentation approaches. \citeauthor{rundo2019novel} \cite{rundo2019novel} apply the MedGA method \cite{rundo2019medga} to find optimal values for normalising the input image. % \citeauthor{cuevas2010novel} \cite{cuevas2010novel} propose a DE method to optimise the parameters of the Gaussian functions of each class based on the histogram of the image. Multiple threshold values are then calculated by computing the overall probability error for two neighbouring Gaussian functions. 
%\citeauthor{khorram2019new} \cite{khorram2019new} propose an ACO method to search for the path from one pixel to neighbouring pixels by using texture features as heuristic information for brain image segmentation. The segmentation results are obtained by comparing the homogeneity values of pixels with a pre-defined threshold value. 

The problem of optimising parameters or threshold values can be formulated as a many-objective optimisation problem based on different objective functions. \citeauthor{abd2019many} \cite{abd2019many} propose a many-objective optimisation (MaOP) method based on Knee point driven evolutionary algorithm (KnEA) to search for threshold values by simultaneously optimising seven objective functions, e.g., Otsu measure, Kapur's entropy, fuzzy entropy, cross-entropy, etc. This method outperforms four MaOP methods on six images. %\citeauthor{xing2021many} \cite{xing2021many} propose a MaOP method based on the boost marine predators algorithm (BMPA) by optimising nine Kapur entropy functions, which are calculated separately using all three channels of the colour images and their combinations.

%\emph{\textbf{Discussions:}}
To sum up, most EC-based methods aim to find optimal threshold values based on Otsu. The most popular methods are based on DE and PSO due to the real-value representation that matches the problem well. The results show that the EC-based methods typically outperform the basic threshold-based image segmentation methods. In addition, different population initialisation and updating strategies and entropy-based fitness functions are often used in these EC methods to improve their performance. EMO algorithms have also shown a promise in handling multiple objectives in threshold-based image segmentation systems. 

\subsection{Region-based Image Segmentation Methods}
Region-based image segmentation methods typically find regions with similar pixel values that satisfy predefined rules by preserving the spatial relationships between pixels. There are two types of region-based methods, i.e., region-growing methods and region-merging methods. A region-growing method is a bottom-up approach that uses some pixels as the seed pixels and grows the region by checking neighbouring pixels. If the neighbouring pixels meet predefined rules, they will be grouped into the same region as the seed pixel. The region merging method, instead, is a top-down method that splits an image into small regions and then merges these regions according to predefined rules. EC techniques have been used to optimise both region-growing and region-merging methods. 

In the region growing-based image segmentation methods, EC techniques have been used to find optimal initial seed pixels. \citeauthor{xie2013automatic} \cite{xie2013automatic} apply GA to select seed pixels by optimising between-class variance and use the self-generating neural forest to generate rules for growing regions. A new strategy is used to adaptively determine the number of seed pixels. This method achieves better performance than other clustering-based image segmentation methods. \citeauthor{gong2017differential} \cite{gong2017differential} propose a DE method with a fitness function integrating within-superpixel error, boundary gradient, and a regularisation term. This method finds a number of seeds and then grows superpixels/regions from the seeds by using DE to find the optimal differential lengths along the vertical and horizontal directions to obtain optimal seeds. Nearest-neighbour is employed to group pixels into small regions. This method shows the powerful global search ability of DE. %\citeauthor{mabrouk2019immune} \cite{mabrouk2019immune} develop the immune system programming (ISP) algorithm to evolve tree-based models as threshold functions in region growing-based image segmentation. The proposed method is a supervised learning method that evaluates the fitness of individuals using labelled pixels. The constructed tree-based models are similar to those evolved by GP, but ISP uses a different evolutionary process. This method shows good performance on several magnetic resonance and computed tomography images. 

In region merging-based segmentation, EC techniques have been used to evolve rules or select regions for merging. \citeauthor{lee2012saliency} \cite{lee2012saliency} propose a PSO-based method for colour image segmentation based on saliency maps. This method consists of four steps, i.e., colour quantisation, feature extraction, small-region elimination, and region merging. PSO is then used to optimise two threshold values for determining the difference between two regions based on the features extracted for region merging. This method provides better colour image segmentation results than four other algorithms. \citeauthor{rogai2016metaheuristics} \cite{rogai2016metaheuristics} apply GA with a real-value encoding to optimise the parameters for a fuzzy logic information fusion method that can score regions produced by a segmentation step and merge those regions to form the final segmentation results. In addition, an ACO-based method is proposed to search for the fuzzy rules for region selection. Both methods provide satisfactory results, ACO being more efficient than GA. \citeauthor{liu2021multiobjective} \cite{liu2021multiobjective} propose an EMO method to merge a set of regions by simultaneously optimising colour similarity and texture similarity. 

To sum up, in both region growing and region merging-based methods, EC techniques have been developed and employed in multiple ways, i.e., selecting the seed pixels, optimising parameters, and evolving region-growing or region-merging rules. These methods use different representations and problem-dependent fitness functions. The promising results demonstrate that EC methods are very flexible and easy to adapt to the problem. However, region-based image segmentation methods have not yet thoroughly explored both the potential of EC methods to improve the search mechanisms and possible hybridisations with other existing methods.

%\emph{\textbf{Discussions:}}

\subsection{Clustering-based Image Segmentation Methods}
Clustering-based methods aim to group pixels into different clusters that have low inter-cluster and high intra-cluster similarity to achieve image segmentation. Clustering-based methods are a kind of region-based image segmentation methods using techniques such as K-means clustering, meanshift, and fuzzy C-means, to obtain homogeneous regions. In these methods, prior knowledge is often needed to set the number of clusters. In addition, the initial cluster centroids are often selected randomly, which affects the performance and the stability of the method. To address these limitations, EC techniques have been proposed to improve the clustering performance in image segmentation by finding the optimal number of clusters and/or the optimal initial cluster centres.

 \citeauthor{khan2015modified} \cite{khan2015modified} propose a DE method to search for the optimal number of clusters in a spatial fuzzy C-mean clustering method for image segmentation. The DE method is improved by introducing a mutation operator based on an archive and an opposition-based population initialisation strategy. The average ratio between fuzzy overlap and fuzzy separation is used as a cluster validity index to be optimised by DE. This method outperforms several image segmentation methods.

EC methods have been used to find good initial cluster centroids for image segmentation. \citeauthor{khan2015genetic} \cite{khan2015genetic} apply a self-organising map to generate the initial set of clusters and propose a GA method with an opposition-based population initialisation strategy to evolve clusters in the fuzzy C-means method for colour image segmentation. In GA, a chromosome is encoded by a string of 3$\times k$ real values, representing $k$ cluster centres, each represented by three colour channels. The fuzzy C-means clustering method groups all pixels into the corresponding clusters, and the cluster compactness and separation are optimised by the GA. This method achieves better performance than eight other methods in terms of qualitative and quantitative results. %\citeauthor{zhao2019alternate} \cite{zhao2019alternate} propose a PSO method with a multiscale position updating strategy to optimise the fuzzifiers and initial cluster centroids in Type-2 intuitionistic fuzzy c-means clustering \cite{hwang2007uncertain} for colour image segmentation. 
\citeauthor{singh2019sunflower} \cite{singh2019sunflower} apply PSO to evolve cluster centres for image segmentation in a sunflower leaf disease detection system. \citeauthor{zhao2018noise} \cite{zhao2018noise} propose an EMO method for multi-objective clustering-based image segmentation. To fully handle noise in images, a noise-robust intuitionistic fuzzy set is defined and the EMO method is used to find the optimal intuitionistic fuzzy representation of the cluster centres while simultaneously optimising the compactness and separation functions. \citeauthor{zhao2020semisupervised} \cite{zhao2020semisupervised} propose a method that builds a Kriging model to predict the fitness values in an EMO for finding cluster centres by simultaneously optimising fuzzy compactness and separation. The results show that this method achieves competitive performance and short computation time. %\citeauthor{zhang2019unsupervised} \cite{zhang2019unsupervised} propose an EMO-based sampling method and an EMO-based clustering method for image segmentation. The former searches for a small number of pixels containing most of the information in the image and the latter evolves cluster centres, activation indexes, weights, and control parameters. The segmentation results are further enhanced using a label correction method based on entropy and local spatial information.%TThis method is further extended in \cite{jiao2020two} by proposing a two-stage method for noisy image segmentation. 

EC methods have been used to evolve both the number of clusters and the initial cluster centroids. \citeauthor{das2009automatic} \cite{das2009automatic} apply a DE method to automatically find the optimal number of clusters and the cluster centres in the fuzzy c-means method for image segmentation. A DE individual is encoded by a vector of $c_{max}$ and $d\times c_{max}$ real values, denoting the selection of clusters from a predefined maximal number $c_{max}$ of clusters and their centres, represented by $d$-dimensional vectors. The fitness function is based on three different cluster validity indices. This method achieves better performance than the GA-based method and the fuzzy c-means method. \citeauthor{das2010kernel} \cite{das2010kernel} further improve the performance of the DE method in \cite{das2009automatic} by developing a kernel-based cluster validity index, i.e., the kernelised Xie-Beni index, as the fitness function and a new neighbourhood topology. This method achieves better results than the GA-based and DE-based methods. %\citeauthor{veenman2003cellular} \cite{veenman2003cellular} propose a cellular coevolutionary algorithm (CCA) to optimise the cluster models using a hard constraint on intra-cluster variability to merge clusters found by CCA and a distributed learning manner. This method addresses the scale problem (i.e., the number of clusters) and the noise problem in image segmentation. 
%\citeauthor{farshi2020multimodal} \cite{farshi2020multimodal} propose a multi-model PSO method with a local search operator to find the number of clusters and the cluster centres (i.e. peak locations) based on the three-dimensional colour image histograms and use the Euclidean distance to associate pixels to each cluster centre. %Instead of encoding the number of clusters in particles, this method defines a parameter representing a distance limit for eliminating clusters in the local search step. 
%This method obtains desirable performance on colour images.

%Different from the above methods, ACO-based methods generate the pheromone matrix of an image based on which a clustering algorithm generates the segmentation. \citeauthor{shahdoosti2019mri} \cite{shahdoosti2019mri} apply ACO to find spatial regions of images based on the colour information of a pseudo-colour image using a clustering method. The segmentation map and the fusion map generated by ensemble empirical model decomposition are combined using majority voting. This method is applied to medical image segmentation with achieving promising performance.

%\citeauthor{han2007improved} \cite{han2007improved} develop an ACO method with new initialisation method and a new heuristic function to allocate ants into different clusters. This method achieves better results than Sobel, Canny and Watershed algorithms. 

To sum up, EC methods, including GAs, PSO, DE, and EMO, are popular for optimising clustering-based image segmentation methods. These methods optimise the number of clusters, initial cluster centres, or both. In these methods, a real-value vector-based representation is typically used to represent the solutions. In the segmentation system, fuzzy C-mean clustering is often used. The EC-based methods typically achieve better performance than traditional clustering methods, and avoid human intervention in setting the number of clusters and the cluster centres. 

%However, the potential of EC methods can be further explored on more complex images such as satellite images. 

%\citeauthor{deepa2019global} \cite{deepa2019global} propose a hybrid method based on biotic cross-pollination algorithm (GBCPA) and ES to search for the cluster centres for real image segmentation. 

%\emph{\textbf{Discussions:}}

\subsection{Classification-based Image Segmentation Methods}
Classification-based methods typically build models or classifiers that are able to associate each pixel in an image with different classes to achieve image segmentation. Unlike the majority of the aforementioned image segmentation methods, classification-based methods are supervised methods that need a set of training samples to learn the classifier models. In recent years, classification-based methods including NNs and GP have been very popular for image segmentation because of their powerful learning ability and excellent performance on complex images. EC-based image segmentation methods can be broadly classified into two types, i.e., using EC (mainly GP) to automatically evolve/construct classifiers/models and using EC to optimise existing classification algorithms. 

GP is the main method that automatically evolves models/classifiers for image segmentation without relying on other segmentation or learning methods. Early works on GP for image segmentation can be found in \cite{poli1996genetic, song2008texture, liang2014image}. 
%\citeauthor{song2008texture} \cite{song2008texture} propose a GP method using raw pixel values as terminals to automatically evolve classifiers for texture image segmentation. This method is very fast and the evolved classifiers are well interpretable. The results show that the GP method can handle complex shapes and produce relatively smooth boundaries. In recent years, GP has been widely applied for figure-ground image segmentation, which is a special case of image segmentation that divides images into the foreground and background regions. %\citeauthor{singh2009genetic} \cite{singh2009genetic} propose a GP method with 20 primitive operators to evolve solutions for image segmentation. This method maximises the classification accuracy to search for the best solutions. 
\citeauthor{liang2015supervised} \cite{liang2015supervised} propose several GP methods to evolve classifiers from a set of features and assign each pixel to the foreground or the background. In \cite{liang2015supervised}, a GP method using raw pixel values and pixel statistics as inputs is developed to evolve classifiers for segmentation. This method achieves better results than four traditional methods including thresholding, region growing, clustering, and active contour models. \citeauthor{liang2017genetic} \cite{liang2017genetic} propose a GP method using seven types of features including pixel values, histogram statistics, texture features, Fourier power spectrum, and Gabor features as the terminal set. The GP method using pixel values achieves the best results on texture images while using the Gabor features achieves the best results on two other datasets. The results also show that the GP method evolves interpretable models/programs with implicit feature selection. \citeauthor{liang2019figure} \cite{liang2019figure} propose multi-objective GP methods to evolve solutions with a good balance between the program/solution complexity and the classification accuracy for figure-ground image segmentation. \citeauthor{liang2020evolving} \cite{liang2020evolving} also propose the StronglyGP method featuring an image processing layer, an image segmentation layer and a post-processing layer to automatically evolve image segmentation solutions consisting of image operators. The StronglyGP method achieves better performance than three other GP variants that use a co-evolution mechanism and a two-stage learning strategy to evolve combinations of operators in each layer. %\citeauthor{liang2020genetic} \cite{liang2020genetic} propose a GP method with a single-tree representation to construct multiple features for image segmentation. Unlike traditional GP, this method concatenates all the outputs of the internal nodes of a GP tree as output features. Different classification algorithms such as k-nearest neighbour and decision tree are also investigated to achieve image segmentation.

EC techniques have been applied to optimising neural network (NN)-based image segmentation methods. The NN-based methods, particularly convolutional neural networks (CNNs), have become popular for image segmentation in recent years. However, these methods have a number of limitations such as having a large number of parameters and requiring rich domain knowledge to design the architectures of NNs. EC techniques can address these limitations by automatically searching for the best NN models for a specific task. \citeauthor{zhou2019evolutionary} \cite{zhou2019evolutionary} propose the ECDNN method to find less important filters and prune these filters for compression of deep NNs (DNNs) for biomedical image segmentation. This method optimises the loss of DNNs and the number of parameters simultaneously. The results show that ECDNN can improve the segmentation performance by using more efficient DNNs. \citeauthor{hassanzadeh2020evolutionary} \cite{hassanzadeh2020evolutionary} propose an evolutionary DenseRes model using GA to automatically search for U-Net architectures based on Dense and Residual blocks. %This method optimises 14 parameters of the U-Net architectures such as the number of blocks, number of convolutional layers, type of activation functions, etc. 
This method outperforms several manually or automatically designed U-Nets on six datasets for medical image segmentation. \citeauthor{hassanzadeh2020evou} \cite{hassanzadeh2020evou} propose the EvoU-Net method using GA to evolve small networks based on U-Net for medical image segmentation. EvoU-Net outperforms U-Net and AdaResU-Net using significantly smaller models. \citeauthor{wei2021genetic} \cite{wei2021genetic} propose the Genetic U-Net method using GA to automatically design U-Net based on several building blocks such as ResNet and DenseNet blocks for retinal vessel segmentation. \citeauthor{lima2021automatic} \cite{lima2021automatic} design a grammar to define the building blocks of U-Net and propose the dynamic structured grammatical evolution to automatically evolve U-Nets for edge detection and image segmentation. 

%\emph{\textbf{Discussions:}}

To sum up, the classification-based methods typically use GP to automatically evolve models/classifiers and use EC-based methods to optimise CNN-based models. The GP-based methods typically find easily interpretable tree-based models consisting of functions and terminals/features to group pixels into different classes. The EC-based CNN methods aim to find promising CNNs to achieve an effective image segmentation. These methods address the limitations of existing CNN methods. Unlike the threshold-, region-, and clustering-based image segmentation methods, classification-based methods are supervised learning methods that use a set of training images to learn models. These methods are often more effective for complex images including natural images and medical images compared to other types of methods {\color{blue}\cite{muangkote2017rr, zhao2020semisupervised}}. EC methods have shown potential in deriving classification-based methods for image segmentation, but there is still room for research, including exploring new solution representations, new applications, computationally-efficient fitness evaluations, and multi-objective search mechanisms.

\section{Image Feature Analysis} \label{featureanalysis}
\begin{figure}
	\centering
	\includegraphics[width=\linewidth]{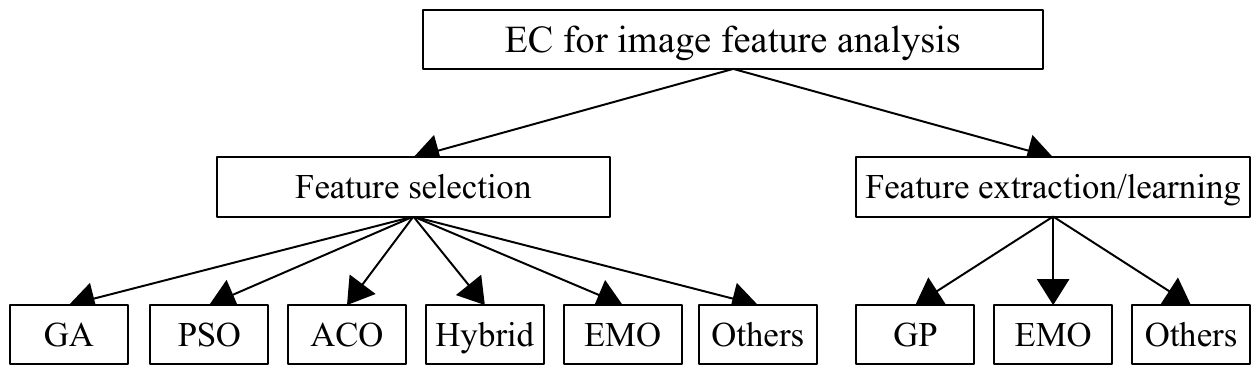}
	% \vspace{-6mm}
	\caption{A brief summary of existing EC methods for image feature analysis.}
	\label{fig:imagefeatureanalysis}
		% \vspace{-4mm}
\end{figure}

Image feature analysis is an important process in many image-related tasks. Images are typically represented by rasters of raw pixels that may contain both meaningful and useless information. Image features are often used to achieve a compact image representation containing as much useful information as possible. Image feature analysis focuses mainly on analysing the meaningful information/features of images. It typically includes feature selection, feature extraction, feature construction or a more general step, i.e., feature learning. The goal of image feature analysis is to improve the performance of a CV task, such as classification and recognition. This section reviews and discusses works related to image feature analysis. A summary of existing EC methods for image feature analysis is shown in Fig. \ref{fig:imagefeatureanalysis}.

%\begin{table}[htbp]
%	%	\vspace{-4mm}
%	%	\footnotesize
%	%	\setlength{\tabcolsep}{0.6em} % for the horizontal padding
%	%\renewcommand{\arraystretch}{1.13}
%	\caption{Summary of EC approaches to image feature selection}
%	%	\vspace{-4mm}
%	\begin{center}
%		\begin{tabular}{p{0.135\linewidth}|p{0.33 \linewidth}}
%			\hline 
%			Categories&References \\
%			\hline
%			GA & 
%			\cite{bhanu2003genetic, lu2008feature, li2010dynamic, si2009evolutionary, chatterjee2011genetic, lin2014study, alirezazadeh2015genetic, kirar2020combination}
%			\\
%			Others& \\ \hline			
%			
%		\end{tabular}	
%		\label{table:FeatureDetection}
%	\end{center}
%	%	\vspace{-6mm}
%\end{table}
%

%\emph{\textbf{Summary}}. 
\subsection{Feature Selection} \label{feature_selection}
Typically, different types of features, such as colour, histogram, shape, and texture features, can be extracted from images using different manually designed image descriptors or automatically learned descriptors. To represent images with sufficient information, a large number of features are usually desirable. However, some of those features can be redundant or irrelevant for the task at hand. Feature selection that selects the most relevant features from a set of pre-extracted ones \cite{xue2015survey} is a necessary step in image analysis. Feature selection becomes challenging when the number of features is large and there exist  complex interactions between features. With a population-based search mechanism, EC methods including GA, ACO, PSO, hybrid methods, and EMO have been widely used to select features for image-related tasks including image annotation, image classification, and image retrieval. 

%An image analysis task often include steps of image preprocessing, feature extraction, feature selection, and regression/classification/detection/recognition. 

%Feature selection is a middle step in this system. 

%There are three key questions to apply EC methods for feature selection, i.e., how to encode the individuals to represent selected feature subsets, how to search for potential feature subsets in the large search space, and how to evaluate the selected feature subsets \cite{xue2015survey}. To address these questions, different EC methods including GAs, ACO, PSO, DE, GP, hybrid methods, and EMO have been developed. %These methods perform an optimisation role by selecting optimal features to solve a task. 

\subsubsection{GA-based Methods}
With a binary string-based encoding/representation, GAs are naturally suitable for feature selection. Specifically, a 1 in a chromosome represents selecting the corresponding feature while 0 represents not selecting it. %are commonly used for feature selection in image analysis. %\citeauthor{bhanu2003genetic} \cite{bhanu2003genetic} apply GA to select features with a new fitness function based on within-class and between-class distance for target detection in SAR images. The results show the fitness function based on distance performs better than the other three fitness functions that measure accuracy and/or feature selection ratio. 
%\citeauthor{lu2008feature} \cite{lu2008feature} investigate GA with a hybrid encoding, i.e., both string-based and real number-based, to achieve feature selection and weighting for image annotation. This method uses KNN to evaluate the selected or weighted features. The results show that GAs with a hybrid encoding achieve better performance than GAs relying on a purely string-based or real number-based encoding. %Based on the same encoding strategies, \citeauthor{li2010dynamic} \cite{li2010dynamic} develop a GA-based feature selection method with an Adaboost learning system to construct ensembles for image annotation. 
%\citeauthor{si2009evolutionary} \cite{si2009evolutionary} propose a GA method to select features that are effective in two different domains, where the selected features are evaluated using an integrated function for cross-domain learning. The results show that GA is effective for dimensionality reduction in cross-domain learning for image annotation and face recognition. 
%\citeauthor{chatterjee2011genetic} \cite{chatterjee2011genetic} develop a GA method to select image features for measuring the quality parameters of materials. The overall process includes image acquisition, image segmentation, feature extraction, feature reduction, and classification/regression. In the experiments reported, GA could reduce the number of features from 189 to 39 and achieve better performance than an existing method on images sampled from different locations. 
\citeauthor{lin2014study} \cite{lin2014study} propose a GA method to select colour and texture features for texture and object image classification. This method achieves better performance than traditional feature selection methods and other existing methods on four datasets. \citeauthor{alirezazadeh2015genetic} \cite{alirezazadeh2015genetic} improve the GA-based feature selection method by introducing new crossover and mutation operators and using different types of hand-crafted features, including linear binary patterns (LBP) and histogram of oriented gradients (HOG) features, for kinship verification. This method achieves better performance than several metric learning-based methods and feature learning-based methods on two datasets. %\citeauthor{mohammed2018real} \cite{mohammed2018real} propose a GA method to select nasopharyngeal carcinoma features to detect cancerous tissues from endoscopic images. This method segments the original images into different small regions, extracts Haar features from those regions, selects a small number of features using GA, and classifies these regions into different classes using NNs. This method achieves high accuracy on one dataset. 
\citeauthor{hemanth2019modified} \cite{hemanth2019modified} propose three GA variants with different offspring generation strategies in the crossover and mutation operators to select texture features for classifying tumour images. \citeauthor{kirar2020combination} \cite{kirar2020combination} propose a two-stage feature selection method where the Quantum GA method is employed to further reduce the number of features selected by a graph-theoretic filter method for motor imagery classification.

\subsubsection{PSO-based Methods} In recent years, PSO has become a popular image feature selection method. %\citeauthor{broilo2010stochastic} \cite{broilo2010stochastic} apply PSO to weight the features in a content-based image retrieval system based on relevance feedback. This method extracts different types of image features and applies PSO to find optimal feature weights in a distance-based similarity function to improve image retrieval results. The user's feedback is employed to update the relevant and irrelevant image sets, which are used to further guide the PSO search. PSO can be extended to retrieving multiple relevant images by using multiple small populations, where each population searches for one relevant image. This method achieves better performance than traditional deterministic methods. 
%\citeauthor{aneesh2012optimal} \cite{aneesh2012optimal} propose an accelerated binary PSO method to select face features for face recognition. In this method, image preprocessing is conducted and discrete cosine transformation features are extracted. The binary PSO is improved by using a new position update rule to perform feature selection, while KNN is used as a classifier. This method improves the classification accuracy on two datasets while achieving a 50\% reduction in the number of features. 
%\citeauthor{jin2015automatic} \cite{jin2015automatic} propose an improved quantum PSO for feature selection in image annotation. In this method, colour, edge histogram, texture, colour layout, and colour structure features are extracted from images and concatenated to form a 250-D feature vector. The improved quantum PSO method with a binary encoding is used to perform feature selection. A boosting ensemble method with three types of classifiers is used to perform classification using the selected features. The results show that this method improves the image annotation performance. %\citeauthor{khan2018face} \cite{khan2018face} propose the binary PSO method to select features for face recognition. 
\citeauthor{naeini2018particle} \cite{naeini2018particle} develop a binary PSO-based method to select features from pre-extracted spectral, textural, and structural features for object-based classification in satellite imagery. A minimum-distance classifier is used to perform classification. This method achieves better performance than using all features and the feature selection methods based on GA and ABC. \citeauthor{tan2018intelligent} \cite{tan2018intelligent} propose a PSO method with two subswarms and corresponding search strategies to select features for skin cancer detection. This method outperforms a large number of EC-based and non-EC-based feature selection methods. 
\citeauthor{tan2019intelligent} \cite{tan2019intelligent} investigate different strategies such as adaptive and random acceleration coefficients, sub-dimension search and re-initialisation in PSO and propose two enhanced PSO methods for feature selection. An ensemble of classifiers is built to achieve effective skin cancer diagnosis. \citeauthor{kavuran2021sem} \cite{kavuran2021sem} has developed a PSO method to select features extracted from pre-trained deep CNN models (i.e., AlexNet and ResNet-50) for the classification of scanning electron microscope images. This method achieves good performance by finding a small number of features. However, this task is computationally expensive, so a small population size and a small number of iterations are used for PSO, which may lead to finding local optima.

\subsubsection{ACO-based Methods}
ACO has also been widely used for image feature selection. 
\citeauthor{rashno2017effective} \cite{rashno2017effective} present two feature selection methods based on ACO and extreme learning machine using wavelet and colour features for pixel classification of Mars images. The first method selects a feature subset for all the pixel classes, while the second method selects feature subsets for each pixel class. The results show that both methods are effective for reducing the running time, while the second method achieves higher accuracy. \citeauthor{sweetlin2017feature} \cite{sweetlin2017feature} use ACO to select features for detecting lung disorders from computed tomography images. This method detects regions of interest (ROI) from images and extracts texture and geometric features from each ROI. The ACO method with a tandem run recruitment strategy is used to search for promising paths towards important features. This method achieves better performance than using all the original features. %\citeauthor{alddd2018image} \cite{alddd2018image} propose an ACO method to select features for segmenting very high spatial-resolution aerial images. 
\citeauthor{devarajan2020metaheuristic} \cite{devarajan2020metaheuristic} propose an ACO method to select features for a NN classifier for medical image segmentation. This method achieves better performance than Bayesian net and NN. 

\subsubsection{EMO-based Methods} Feature selection often involves two objectives: minimising the number of selected features and maximising the performance measure, e.g., classification accuracy, which are potentially conflicting with each other. EMO-based methods have been adopted for multi-object feature selection in image analysis. \citeauthor{mlakar2017multi} \cite{mlakar2017multi} present a multi-objective DE method to select features for facial expression classification by minimising the number of features and maximising the classification accuracy. This method extracts HOG features from small facial regions of different sizes. The authors investigate two selection strategies, i.e., selecting emotion-specific features and the most discriminative features for all emotions. The effectiveness of these methods is verified on three face image datasets and compared with using all features and other existing methods. \citeauthor{liang2017image} \cite{liang2017image} propose a single-objective GP method and two multi-objective GP methods to automatically select features from a set of edge, colour, and statistical features for image segmentation. The results show that the multi-objective GP methods achieve better performance using a smaller number of features.

\subsubsection{Hybrid Methods}
The hybridisation of EC methods has been proposed for image feature selection. \citeauthor{ghamisi2014feature} \cite{ghamisi2014feature} propose a hybrid method based on PSO and GA for feature selection in hyperspectral image analysis. This method achieves better performance than individual GA and PSO methods on two datasets. \citeauthor{thangavel2014soft} \cite{thangavel2014soft} present hybrid methods based on rough sets, GA and ACO to classify cancer images. The hybridisation of ACO and rough sets achieves better classification accuracy than the other hybrid methods. However, the total number of original features they consider is 22, which is quite small.  %\citeauthor{jothi2016hybrid} \cite{jothi2016hybrid} propose a hybrid method based on rough sets and the firefly algorithm to select features for MRI brain tumour image classification. This method achieves good performance in removing redundant features. 
\citeauthor{mistry2016micro} \cite{mistry2016micro} embed PSO with the concepts of micro-GA to select features for facial emotion recognition. This method outperforms PSO variants, classic GA and PSO, and other methods.

\subsubsection{Other Methods} Other EC methods such as DE and GP have also been studied for image feature selection. \citeauthor{ghosh2013self} \cite{ghosh2013self} develop a self-adaptive DE (SADE) method to select features for hyperspectral image classification. This method uses ReliefF to reduce the redundant features and a fuzzy KNN to evaluate the effectiveness of features. SADE achieves better performance than four other EC methods, i.e., GA, ACO, DE, and a combination of ACO and DE, in terms of accuracy and kappa coefficient. %\citeauthor{srinivas2018automatic} \cite{srinivas2018automatic} propose a DE-based feature selection method for image clustering based on GLCM features. This method achieves better performance than a GA-based method. 
GP has implicit feature selection properties since the leaf nodes of the GP trees/programs represent the selected features. \citeauthor{ain2017genetic} \cite{ain2017genetic} propose a GP-based feature selection method for skin cancer classification. This method achieves better performance than traditional methods.

To sum up, many EC-based methods including GA, ACO, PSO, EMO, DE, and GP have been developed to select image features for dealing with tasks such as image classification and image segmentation. In those tasks, the image features are often manually extracted and the number of features is often not too large. Therefore, most existing methods deal with a small number of features, e.g., less than 300, thus a moderately-sized set. Complex feature interactions can be handled by using different solution representations, search mechanisms and feature subset evaluation measures in different EC methods. This leads to using different types of EC methods to select image features for different tasks, e.g., image classification and image segmentation. In addition, feature selection is a multi-objective optimisation problem whose goals are maximising the performance on the task and minimising the number of selected features. EMO methods have shown a big potential in dealing with this problem. However, not so many works exist on EMO for feature selection in image analysis. Future research directions can focus on using EC to solve large-scale feature selection problems and multi-objective feature selection problems.

%pros: Reduce dimensionality, improve classification performance, improve interpretability, suitable for different types of image-related tasks that need features, low computation cost, 
%cons: require domain knowledge to extract image features in advance, the quality of preextracted features is very important,
%current EC methods work on very low-dimensional data, 
%very few EMO methods work on image data

\subsection{Feature Extraction and Learning}\label{feature_extraction}
Feature extraction consists in extracting informative features from raw images to solve a task. Feature construction consists in building high-level features as compositions of the original ones. Unlike feature selection, feature extraction and construction can therefore generate new features to solve a task \cite{bi2021gpimage}. When a learning algorithm is used to learn features by performing feature selection, extraction or construction from the original data, the task is also termed {\em representation learning} or {\em feature learning}. EC methods have been used for feature extraction and construction in image analysis. Existing methods focus mainly on optimising existing feature extraction methods, constructing features from pre-extracted features, or automatically learning models/solutions from scratch. 

\subsubsection{Optimisation of Existing Feature Extraction Methods} Several EC methods have been developed to optimise the parameters or components of existing image feature extraction methods. \citeauthor{albukhanajer2014evolutionary} \cite{albukhanajer2014evolutionary} propose a method based on NSGA-II to optimise three functionals in the trace transform and extract effective and robust features that simultaneously optimise two objectives: minimising the within-class distance and maximising the between-class distance. The results show that this method can extract effective and noise-invariant features for image classification. \citeauthor{gong2015multiobjective} \cite{gong2015multiobjective} introduce an EMO method to optimise auto-encoder biases for sparse feature learning from images. This method simultaneously optimises two objectives, i.e., the reconstruction error and the hidden units' sparsity, to learn useful sparse features from images. \citeauthor{albukhanajer2017classifier} \cite{albukhanajer2017classifier} propose an EMO method to find multiple Pareto image features based on the trace transform with a good tradeoff between two different objectives (the within-class variance and between-class variance) and build an ensemble for object classification.

%\cite{shah2017evolutionary} develop a GA method for feature learning from three-dimensional images. This method used 

%(add a figure to show the overall process)
\subsubsection{Automatic Evolution of Models/Descriptors for Feature Extraction}
GP-based methods have been widely used for feature extraction and learning from raw images by automatically evolving solutions from scratch. In other words, the input of a GP system is an image, while the output is a single feature or a set of image features. Early works on using GP for evolving descriptors can be found in \cite{krawiec2005visual, krawiec2007visual, perez2009evolutionary, perez2013genetic} .

Texture is an important image property considered in EC-based applications. \citeauthor{al2016automatically} \cite{al2016automatically} propose a GP-based method to automatically evolve texture descriptors to extract rotation-invariant texture features from images. This method uses a tree-based representation, where the leaf nodes are pixel statistics computed from a small region of the image and internal nodes are arithmetic operators. The root node of the tree is a special one having a predefined number of child nodes to output a binary vector. The way of using the evolved GP trees for texture description is similar to LBP, i.e., transforming the pixel values by applying a rule (i.e., the GP tree) and generating a histogram as features. The GP-evolved descriptor improves the classification performance over the traditional texture descriptors, such as LBP and its variants, on several texture classification tasks. However, the descriptors can only generate a fixed number of texture features. Therefore, in \cite{al2017keypoints} an improved GP method is developed to extract a flexible number of texture features. Specifically, the method uses a root node accepting a flexible number of child nodes. This method shows its superiority in classifying several datasets in comparison to other texture descriptors. \citeauthor{al2020automatically} \cite{al2020automatically} use a multi-tree GP method instead of a special root node to automatically evolve texture descriptors for classification, where only two instances per class are used in the learning process. 

\citeauthor{rodriguez2018structurally} \cite{rodriguez2018structurally} propose a structured layered GP method for representation learning. This method uses two GP forests, i.e., an encoding forest and a decoding forest. The representation of the image data is learned from the encoding forest. \citeauthor{rodriguez2019evolving} \cite{rodriguez2019evolving} investigate different population dynamics and genetic operators in GP-based autoencoders for representation learning. \citeauthor{rodriguez2020cooperative} \cite{rodriguez2020cooperative} propose a cooperative co-evolutionary GP method to automatically construct features from raw pixels. This method considers co-evolution at genotype, feature, and output levels. The results show that the co-evolution at the output level is the most effective. 

In recent years, many GP methods with image-related operators have been developed to automatically learn/extract different types of image features. \citeauthor{liu2015learning} \cite{liu2015learning} propose a multi-objective GP method with a multi-layer representation to learn spatio-temporal features for action recognition. The results show that the features learned by GP are more effective than manually extracted ones and other machine-learned features for action recognition. \citeauthor{liu2015sequential} \cite{liu2015sequential} present a GP method to learn models that generate a low-dimensional representation for image hashing. This method evolves multiple GP trees to transform features into binary codes by optimising the empirical risk with a boosting strategy on the training set. This method achieves promising results on two large datasets. \citeauthor{bi2018gaussian} \cite{bi2018gaussian} investigate the use of Gaussian filters in GP for image feature learning for classification. In \cite{bi2018genetic} and \cite{bi2020effective}, two GP methods with well-developed image descriptors as functions are proposed to automatically learn global and/or local features for image classification. The methods are very effective in different image classification tasks. \citeauthor{bi2019tevc} \cite{bi2019tevc} propose a GP method with a flexible program structure and many image-related operators to learn various types and numbers of features from raw images for classification. Similar to feature selection, feature extraction can also be formulated as a multi-objective optimisation problem. In \cite{bi2020automatically} and \cite{bi2021multi}, multi-objective GP methods are proposed to automatically learn features for face recognition. These methods simultaneously maximise the classification accuracy and minimise the number of features.

A potential issue with using EC methods for automatic feature extraction is the high computational cost due to a large number of fitness evaluations. \citeauthor{bi2021divide} \cite{bi2021divide} propose a divide-and-conquer GP method using multiple small populations to learn features from small subsets of the original large training set. This method not only reduces the computational cost but also improves the generalisation performance. In \cite{bi2021instance}, an instance selection-based surrogate method is developed to use multiple well-selected subsets of the large training set to perform fitness evaluation in GP for image feature learning. These methods can significantly reduce the training time without affecting, or even improving, their performance. 

To sum up, in recent years, many EC-based methods, particularly GP-based methods, have been developed for image feature extraction and learning for different tasks including texture classification, image classification and action recognition. Compared with traditional methods, the EC-based methods require less domain knowledge and are easier to adapt to different tasks. These methods can find optimal parameter values of existing solutions or automatically evolve effective image descriptors for feature extraction from scratch. Therefore, EC-based feature extraction methods typically improve the performances of the tasks being solved. Image feature extraction can often be treated as the multi-objective problem of maximising the performance of the task and minimising the number of features. However, very few EMO methods have been developed to address this problem. In addition, the high computational cost of using EC for image feature extraction needs further attention in the future.

%\subsection{Others}
 
%\subsection{Optimisation in Specific Solutions/Models}
%\subsection{Section Summary}
%This section discussed related work on EC for image feature analysis, including edge detection, interest point detection, feature selection, feature extraction and learning. 

\section{Image Classification}
\begin{figure}
	\centering
	\includegraphics[width=\linewidth]{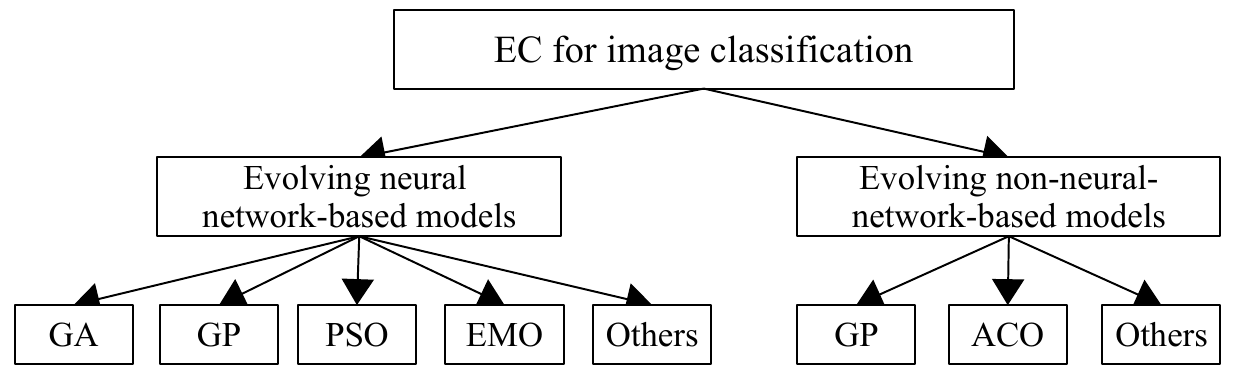}
	\caption{Summary of EC approaches to image classification. }
	\label{fig:imageclassification}
\end{figure}

Image classification is a fundamental and essential task in CV and machine learning. Image classification aims to assign an image to one out of a set of predefined classes based on its content. It has a wide range of applications to many important fields, such as medicine, remote sensing, biology, engineering, and business \cite{bi2021gpimage}. However, it is a difficult task due to a high image variability, caused by scale variations, deformations, occlusions, view-point changes, and rotations. The task becomes even more challenging when the image data are noisy, blurred and/or affected by other kinds of distortion. EC techniques have been widely applied to image classification. Existing EC-based methods can be broadly classified into four types, i.e., EC-based feature analysis for image classification, EC-based NNs for image classification, EC for automatically evolving image classification models from scratch, and others. EC-based feature analysis methods are reviewed and discussed in Section \ref{featureanalysis}. Therefore, this section focuses on the remaining categories, as summarised in Fig.~\ref{fig:imageclassification}.

\subsection{EC for Evolving NNs for Image Classification}
Using EC to evolve NNs has been proposed for many years \cite{yao1999evolving}. This topic is becoming increasingly popular in the deep learning era when deeper NN architectures have popularised many important NN variants and components for image-related tasks, like CNNs. Recently, EC techniques have been successfully used to evolve DNNs, in particular deep CNNs, for image classification. Recent related works can be found in  survey papers \cite{zhou2021survey, liu2021survey}. %In \cite{vargas2021review}, a detailed review of EC encoding schemes for evolving CNNs is presented. 
%\citeauthor{zhou2021survey} \cite{zhou2021survey} review typical work on using EC to automatically construct DNNs. \citeauthor{liu2021survey} \cite{liu2021survey} review and discuss recent works on evolutionary neural architecture search in terms of encoding space, encoding strategy, population updating, fitness evaluation, and applications. 
In this section, we review and discuss representative works focused on image classification. %The existing works are grouped into three important groups, i.e., evolving NN weights and/or architectures, EMO for evolving NNs, and improving efficiency for evolving NNs.

\subsubsection{Evolving NN Weights and/or Architectures}
EAs are the most common methods for evolving DNN weights and architectures. \citeauthor{sun2019evolving} \cite{sun2019evolving} propose the EvoCNN method that uses a variable-length encoding to evolve CNN architectures and connection weights. This method has shown its superiority in comparison with 22 existing methods on nine image classification datasets.  %\citeauthor{anaraki2019magnetic} \cite{anaraki2019magnetic} propose a GA method to automatically design CNNs by searching architectural parameters and training parameters for brain tumour grades classification.
\citeauthor{sun2020automatically} \cite{anaraki2019magnetic} present the CNN-GA method that uses GA with a variable-length representation to automatically evolve CNN architectures based on the idea of skip connections. CNN-GA achieves promising performance in terms of classification accuracy, number of parameters, and computation time. \citeauthor{chen2020evolving} \cite{chen2020evolving} propose a GA method with a variable-length encoding scheme for evolving the architectures of deep convolutional variational autoencoders consisting of four different blocks. The evolved NNs are then trained and applied to image classification achieving performance competitive with nine autoencoder variants. \citeauthor{o2021evolutionary} \cite{o2021evolutionary} develop a GA method to automatically search all components, including feed-forward and DenseNet networks, employing a low-fidelity performance predictor for efficient search. This method can evolve different types of skip-connection structures in CNNs to improve the performance of DenseNet. 
 
\citeauthor{real2019regularized} \cite{real2019regularized} propose the AmoebaNet-A method using a specific mutation operator for neural architecture search (NAS). This method discards the oldest models/individuals in the population and selects the newer models for evolution. This method obtains better results and runs faster than reinforcement learning methods. %\citeauthor{xue2021self} \cite{xue2021self} propose an EA method with an adaptive mutation strategy and a semi-complete binary competition strategy to automatically evolve the architectures of CNNs based on blocks. 
\citeauthor{zhang2022evolutionary} \cite{zhang2022evolutionary} propose an evolutionary one-shot NAS method with partial weight sharing to generate sub-models from the one-shot model. New crossover and mutation operators are developed to facilitate weight sharing. %This method uses pyramidal convolution operations in CNNs. 
This method achieves competitive performance with 26 state-of-the-art algorithms on 10 datasets. \citeauthor{zhang2021adaptive} \cite{zhang2021adaptive} propose the reinforced I-Ching divination EA with a variable-length encoding strategy for NAS. A reinforced operator is employed to improve the evolutionary search efficiency. \citeauthor{li2021automatic} \cite{li2021automatic} consider resource constraints in EC-based NAS methods by developing an adaptive penalty method for fitness evaluation and a selective individual repair operation. The model complexity is set as a constraint in order to find small models with high classification performance. 

Besides GA, other EC methods including GP and PSO have been developed to find optimal architectures and/or parameters for CNNs. \citeauthor{suganuma2017genetic} \cite{suganuma2017genetic} propose the CGP-CNN method with a tree-based representation for NAS based on standard convolutional operators or ResNet blocks. This method achieves high classification performance and low model complexity. \citeauthor{li2019evolving} \cite{li2019evolving} apply quantum-behaved PSO to evolve deep CNNs by introducing a binary encoding.  %\citeauthor{singh2021hybrid} \cite{singh2021hybrid} propose a two-level PSO method with multiple swarms to automatically search for deep CNN architectures and hyper-parameters for image classification. 
\citeauthor{gong2020evolving} \cite{gong2020evolving} propose a co-evolutionary method along with backpropagation to search the parameters of DNN models for image classification. The co-evolution and backpropagation methods are used to learn the weights when one of them cannot optimise the training objective function. In co-evolution, the parameter learning task is decomposed into many small tasks for effective search. This method is better than the traditional parameter learning techniques in DNNs.

\subsubsection{EMO for Evolving NNs}
Evolving NNs can be formulated as a multi-objective optimisation problem that simultaneously optimises multiple potentially conflicting objectives, such as accuracy, model size and computational complexity. EMO methods have been developed to optimise CNNs for image classification. \citeauthor{lu2020multiobjective} \cite{lu2020multiobjective} propose the NSGANetV1 method for NAS by simultaneously optimising two objectives, i.e., the classification accuracy and the number of floating-point operations. This method achieves better performance than many manually designed or automatically evolved CNNs on several datasets. %\citeauthor{wang2021evolutionary} \cite{wang2021evolutionary} propose a multi-objective EC method for NN model compression to simultaneously optimise efficiency and accuracy in image classification. This method outperforms the compared methods on CIFAR-10. 
\citeauthor{lu2021neural} \cite{lu2021neural} propose the NAT method to automatically search subnets of CNNs based on NSGA-III. A surrogate model is built to predict model performance during the evolutionary search. NAT can find multiple subnets with a good trade-off between different objectives that can be easily transferred to solve other problems. This method achieves outstanding performance on 11 datasets. \citeauthor{wen2021two} \cite{wen2021two} develop a two-stage EC-based NAS method that transfers the searched subnetworks from a source task to a target task. The knee-guided multi-objective evolutionary algorithm is used as the search mechanism. The method demonstrates the power of evolutionary NAS in transfer learning. \citeauthor{zhu2019multi} \cite{zhu2019multi} propose an EMO-based method to optimise the NN structures in federated learning. This method optimises multilayer perceptron and CNN models with reduced communication costs. \citeauthor{zhu2021real} \cite{zhu2021real} develop an EMO-based NAS method under the real-time federated learning framework with the goals of maximising model performance and minimising local payload. This method can effectively reduce computational and communication costs by using a double-sampling technique. \citeauthor{wang2019evolving} \cite{wang2019evolving} present a multi-objective PSO-based NAS method, which simultaneously optimises two objectives, i.e., classification accuracy and the number of floating-point operations.

\subsubsection{Computation Efficiency in EC for Evolving NN}
Using EC to optimise NNs is computationally expensive because it requires a large number of fitness evaluations. Therefore, recent research focuses on developing computationally cheap evaluation strategies to improve search efficiency. \citeauthor{zhang2020efficient} \cite{zhang2020efficient} propose an evolutionary NAS method that uses small sampled training data and a node inheritance strategy to reduce the computation cost of fitness evaluation. In \cite{lu2020nsganetv2}, \citeauthor{lu2020nsganetv2} propose NSGANetV2, employing surrogates to predict the performance of CNN models during the evolutionary process. A selection method dubbed adaptive switching is proposed to automatically select one of four surrogate models for fitness prediction. \citeauthor{sun2019surrogate} \cite{sun2019surrogate} propose a surrogate-assisted EC-based method for NAS by building an offline performance predictor based on random forest. This method achieves comparable performance with less computational cost than 18 methods. \citeauthor{wang2021surrogate} \cite{wang2021surrogate} propose a new surrogate model based on SVM using a new training-data sampling method to assist PSO in searching for variable-length CNN blocks. This method achieves competitive performance by reducing training of CNN blocks by 80\% during the search process. %\citeauthor{hu2021multi} \cite{hu2021multi} present a random-weight evaluation method to speed up the fitness evaluations in NSGA for NAS. This method randomly samples the weights for CNNs and only trains the classification layer to reduce the computational cost. This method achieves competitive performance while requiring a few hours for training on a single graphic processing unit (GPU) card on typical benchmarks.

To sum up, EC-based methods including GA, GP, and PSO have shown great potential in optimising CNN-based models for image classification. EC methods have powerful global search ability and a very flexible representation, suitable and effective for optimising complex CNN models. In recent years, this topic has become very popular and the CNNs evolved by EC methods have outperformed many well-known manually-designed CNNs on several image classification datasets including CIFAR10, CIFAR100, and ImageNet. Furthermore, EMO-based methods have been widely used to evolve CNNs by simultaneously optimising multiple objective functions. However, EC-based methods for evolving CNNs have some limitations. One issue is the high computation cost. Therefore, some attempts to improve computational efficiency have been based on methods like surrogates and training-data sampling strategies. However, the potential of EC for evolving NNs has not been fully explored yet on many other types of image classification tasks and NN components or variants. 

%\emph{\textbf{Discussions:}}
\subsection{EC for Evolving Non-NN-based Methods}
%EC techniques have also been developed to automatically evolve non-NN-bsed models/solutions for image classification from scratch. 

\subsubsection{Evolving Classifiers from Image Features}
EC methods have been developed to automatically generate classifiers based on pre-extracted image features. The most commonly used methods are based upon GP. %\citeauthor{parkins2004genetic} \cite{parkins2004genetic} propose the early work of using GP to automatically evolve classifiers with implicit feature selection for classifying the USPS dataset, which is a digit recognition task. Different variations on the selection and genetic operators are introduced. %\citeauthor{agnelli2002image} \cite{agnelli2002image} propose a GP-based method to evolve classifiers based on 12 pre-extracted statistical features for classifying text and picture segments. This method is tested on one dataset and achieves high accuracy. 
\citeauthor{choi2012genetic} \cite{choi2012genetic} apply GP to classify nodules and non-nodules in computed tomography images using pre-extracted 2D and 3D features. This method achieves very high sensitivity. \citeauthor{ryan2015image} \cite{ryan2015image} employ GP to evolve classifiers for detecting breast cancer in digital mammograms obtaining 100\% sensitivity. \citeauthor{ghazouani2021genetic} \cite{ghazouani2021genetic} extract geometric and texture features and propose a GP method to perform classification of facial expression images.

GP is naturally suited for evolving binary classifiers but needs special designs/strategies to perform multi-class image classification. \citeauthor{smart2003classification} \cite{smart2003classification} propose three GP-based classification strategies for multi-class object classification, namely static range selection, centred dynamic range selection, and slotted dynamic range selection, which use different threshold values to determine the class labels according to the outputs of GP trees. \citeauthor{zhang2006using} \cite{zhang2006using} propose a GP method with a fitness function that computes the overlap of the distributions of all possible pairs of classes for multi-class object classification using domain-independent features. The classification decision is based on multiple GP classifiers and normal probability density function. This method performs well on three datasets. %\citeauthor{zhang2007new} \cite{zhang2007new} improves the a new crossover operator in GP for multi-class object classification, where a local hill-climbing search based on node looseness values is used to select good crossover points. %This method has been tested on three object classification datasets.

%Besides GP, ACO has also been used to evolve classifiers. \citeauthor{omkar2007urban} \cite{omkar2007urban} propose two EC-based methods for urban satellite image classification. The first, Ant-Miner, uses ACO to evolve classification rules while the second uses PSO to optimise multilayer perceptrons. These methods achieve high accuracy on a five-class classification problem. 

\subsubsection{Evolving Classifiers from Raw Pixels}
In the most common methods, GP evolves tree-like models from raw pixels to achieve image classification. Multi-tier or multi-layer tree structures have been used in GP to evolve binary classifiers based on raw pixels. %\citeauthor{atkins2011domain} \cite{atkins2011domain} propose a multi-tier GP method that constructs models/trees featuring an image-filtering tier, an image-aggregation tier, and a classification tier. In different tiers, a number of different functions are used for learning models. This method classifies images without requiring any pre-extracted features or external classifiers. 
\citeauthor{al2012two} \cite{al2012two} present a two-tier GP method featuring an aggregation tier and a classification tier: the former focuses on detecting small image regions and extracting features while the latter is the image classifier. This method achieves high accuracy on four datasets while the learned classifiers/models show high interpretability. \citeauthor{lensen2016genetic} \cite{lensen2016genetic} propose the HOG-GP method for simultaneous region detection, feature extraction, feature construction, and image classification. Unlike the above methods, HOG-GP extracts features from the detected regions using the HOG descriptor. The method shows how GP can be used to extract high-level features using well-developed descriptors. \citeauthor{burks2018genetic} \cite{burks2018genetic} combine a two-tier GP architecture with the genetic maker diversity algorithm and apply this method to automatically detecting active tuberculosis sites in X-ray images. %\citeauthor{hazgui2021genetic} \cite{hazgui2021genetic} propose a GP method with tiers performing patch detection, texture feature extraction, and classification. Unlike HOG-GP that only uses HOG for feature extraction, this method extracts HOG and LBP features from the selected regions and achieves better performance than HOG-GP on three texture image datasets. 
\citeauthor{bi2018automatic} \cite{bi2018automatic} propose a multi-layer GP method that adds an image filtering layer in order to extract high-level features for image classification. %In \cite{bi2017automatic}, the GP method uses an image processing layer to process the automatically-detected regions and extracts histogram features for image classification. 
\citeauthor{fan2020region} \cite{fan2020region} propose a GP method to extract features using edge detectors, LBP, and fractal dimension functions for image classification. This method achieves performance competitive with a number of other GP methods on four image datasets. %\citeauthor{evans2018evolutionary} \cite{evans2018evolutionary} investigate the use of convolution operators in GP for feature extraction and binary classification of images.

GP has also been used to evolve end-to-end models for binary and multi-class image classification. \citeauthor{fan2022genetic} \cite{fan2022genetic} propose a GP method that can automatically extract features using image filtering functions and image descriptors and select functions suitable for building image classifiers. This method also uses a new mutation operator that dynamically adjusts the sizes of trees, achieving promising performance on eight datasets. \citeauthor{bi2020genetic} \cite{bi2020genetic} propose the IEGP method that, besides extracting features from raw images, also evolves ensembles of image classifiers built using those features in a single tree. This method achieves better performance than a large number of methods on 13 image datasets. \citeauthor{bi2020evolving} \cite{bi2020evolving} propose the EvoDF method in which GP evolves tree-like models based on image feature-extraction operators and random forest ensembles. This method can evolve shallow or complex models by combining these operators effectively. The EvoDF method achieves better performance than several traditional and existing methods on reference datasets using only a small number of training images.

Other EC methods have been used to optimise classification from raw pixels. \citeauthor{plichoski2021face} \cite{plichoski2021face} develop a DE method to automatically select a subset of predefined image-processing and feature-extraction operators and the parameters of a face-recognition algorithm. This method uses a fixed-length binary-vector encoding in which 1 denotes selecting the corresponding operation and 0 denotes not selecting it. 

\subsubsection{Feature Learning and Emerging Topics}
Most works on feature analysis using EC for image classification have been reviewed in Section \ref{featureanalysis}. This subsection will review more works that are not covered in Section \ref{featureanalysis} and emerging topics. \citeauthor{shao2013feature} \cite{shao2013feature} propose a GP-based method to automatically learn image features that simultaneously minimises the tree size and the classification error. This method uses image filters as functions to construct descriptors and achieves better performance than manually-designed features and other two feature-learning methods on four datasets. \citeauthor{agapitos2015deep} \cite{agapitos2015deep} propose a GP method with a filter bank layer, a transformation layer, an average pooling layer, and a classification layer, similar to deep CNNs, to automatically extract features from raw pixels for classifying digits. %\citeauthor{tian2020evolutionary} \cite{tian2020evolutionary} have developed an EP method to select layers from four famous deep CNN models for feature extraction and then select informative features for image classification. The results show that EP achieves better performance than GA. %\citeauthor{lillywhite2012self} \cite{lillywhite2012self} propose a GA-based method known as evolution-constructed features (ECO) for object classification. In this method, a string represents a region selected from the image and the corresponding operators to process the region. An Adaboost classifier is built using multiple ECO features for effective classification. This method achieves promising results on three datasets. An improved version of ECO also exists \cite{lillywhite2013feature}. 
\citeauthor{olague2017brain} \cite{olague2017brain} propose a brain programming (BP) method that uses a multi-tree representation to automatically evolve a set of descriptors to generate features within a hierarchical structure for classification. The BP method is further investigated in \cite{ibarra2022brain} under the scenario of adversarial attacks. The results show that BP is more robust against adversarial examples than deep CNNs on art media categorization tasks.

GP-based methods have shown great potential in image classification with a small number of training images. \citeauthor{al2016binary} \cite{al2016binary} propose one-shot GP to evolve classifiers directly from images and a compound GP for region detection, which facilitates feature extraction for binary image classification. This method only needs a few training images and achieves better performance than traditional methods. \citeauthor{bi2021dual} \cite{bi2021dual} propose a GP-based method with a dual-tree representation and design an efficient and effective fitness function to improve the generalisation performance of GP for few-shot image classification. This method achieves better performance than popular few-shot learning methods under the 3-shot and 5-shot scenarios on nine image datasets. In \cite{bi2022using}, the poor generalisation caused by using small training data is further addressed by developing a new multi-objective GP method to simultaneously optimise two objectives, i.e., the classification accuracy and a distance measure  between  instances (i.e. distances between instances from the same class and from different different classes). The results show that GP  simultaneously optimising these two objective functions achieves better results than GP optimises only one objective function.

Transfer learning has also been investigated in EC to evolve non-NN-based models. \citeauthor{iqbal2017cross} \cite{iqbal2017cross} transfer the code fragments of GP trees evolved from one source dataset/task to improve the learning performance of GP on a target data/task. This method is focused on texture and object classification. \citeauthor{bi2021learning} \cite{bi2021learning} propose a GP method with a multi-tree representation and a co-evolution learning strategy to learn features for simultaneously solving two image classification tasks. The knowledge of the two tasks is encoded as GP trees and explicitly shared during the learning process; this improves the performance of GP when the training set is small.
%Image classification may turn into an imbalanced classification problem. \citeauthor{acharya2020novel} \cite{acharya2020novel} propose a fitness function based on a new score measure for GP to handle the imbalance problem in classifying positive and negative emotions. 
%The image data may contain different levels of distortions such as noise and low contrast. To address this problem, \citeauthor{bi2021genetic} \cite{bi2021genetic} propose a GP method that can learn features from low-quality images by evolving models using a wide set of operators including region detection, filtering/preprocessing, pooling, feature extraction, and concatenation. This method achieves better and more stable performance than traditional methods and the methods using deep features on low-quality images with different levels of distortions. 

To sum up, evolving non-NN-based models using EC methods is also promising in many image classification tasks. The majority of these methods are based on GP, which typically evolves tree-based models with different types of operators. Compared with the CNNs, the models evolved by GP are typically less complex and more interpretable. In addition, GP has shown a promise in learning effective models for image classification using a small number of training images, while CNNs typically need a large training set. In recent years, some new ideas, such as surrogate functions, transfer learning and few-shot learning, have been investigated in the GP-based methods for image classification. Future research can further explore the above aspects and focus on some less explored directions such as developing GP with deep structures, and investigating more computationally efficient GP methods.

%\citeauthor{ye2018application} \cite{ye2018application} develop a bat algorithm to automatically find the parameters for the tunned mask for texture feature extraction for classification. The results show that this method achieves higher accuracy than PSO and GA. However, this method only optimises 25 dimensions, which is very small.

%\citeauthor{luckehe2018evolutionary} \cite{luckehe2018evolutionary} apply EC to simplify the images that are generated by CNN for lung nodule classification. This method could reduce irrelevant parts of the images and improve the interpretability of the classification system.

%\begin{table}[htbp]
%	%	\vspace{-4mm}
%	%	\footnotesize
%	%	\setlength{\tabcolsep}{0.6em} % for the horizontal padding
%	%\renewcommand{\arraystretch}{1.13}
%	\caption{Summary of EC Approaches to Image Classification}
%	%	\vspace{-4mm}
%	\begin{center}
%		\begin{tabular}{p{0.34 \linewidth}|p{0.36 \linewidth}}
%			\hline 
%		Optimisation in specific solutions/models&Automatic evolving of models/solutions from scratch \\
%			\hline	
%		& \\ \hline			
%		& \\ \hline			
%	& \\ \hline			
%		& \\ \hline			
%		\end{tabular}	
%		\label{table:imageClassification}
%	\end{center}
%	%	\vspace{-6mm}
%\end{table}

%\subsection{EC-based Ensemble Learning for Image Classification}

%To sum up, 

\section{Object Detection}
\begin{figure}
	\centering
%	\vspace{-4mm}
	\includegraphics[width=\linewidth]{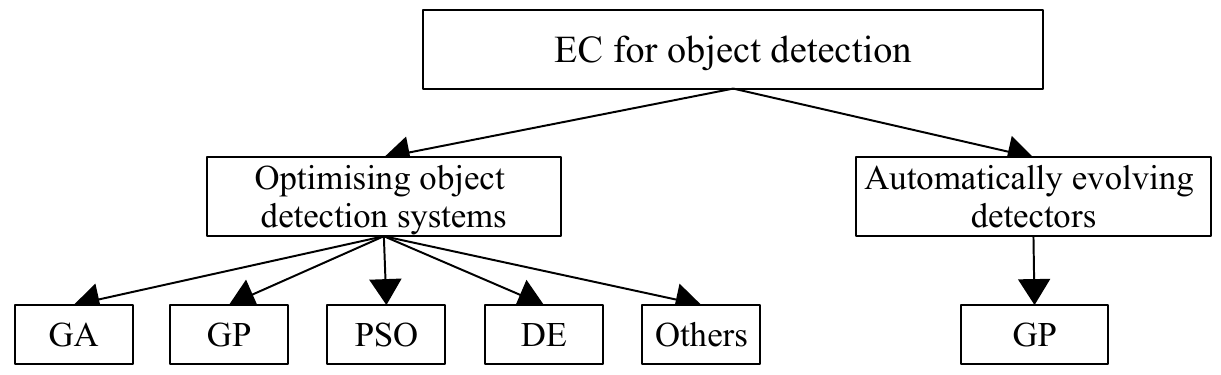}
	% \vspace{-6mm}
	\caption{Taxonomy of EC approaches to object detection. }
	\label{fig:objectdetection}
	% \vspace{-4mm}
\end{figure}

Object detection is a fundamental task in CV and image analysis. It regards not only detecting objects but also locating their position in the images. Thus, object classification and localisation can be considered sub-tasks of object detection, which is therefore more complex than either sub-task. Object detection has many important real-world applications such as detecting humans in self-driving vehicles. EC techniques have been successfully applied to object detection. Existing methods can be broadly classified into two categories, i.e., using EC methods to optimise an object detection system and using EC methods to automatically evolve detectors from scratch. Figure \ref{fig:objectdetection} shows a simple taxonomy of these methods.

% In this section, we will review and discuss existing methods from these two aspects. 

\subsection{Optimising Object Detection Systems} 

In an object detection system, EC methods can be used to optimise features, parameters, positions of objects, etc. \citeauthor{ganesh2014entropy} \cite{ganesh2014entropy} propose an entropic binary PSO method to optimise an entropy map in an ear detection system. A threshold is used to perform classification based on the optimised entropy map. This method achieves promising performance on four face image datasets. \citeauthor{abdel2014efficient} \cite{abdel2014efficient} propose a PSO-based method for eye detection and tracking. In this method, PSO is used to optimise parameters such as the centre point and the scaling factor of the deformable multiple template matching algorithm.
%\citeauthor{chen2015fast} \cite{chen2015fast} propose the HOG-SVM-DE method to achieve fast and effective human detection from images. In this method, DE is used to search for the best location and window size of a region that contains objects. The HOG features are extracted from the detected region and an SVM classifier is used to predict the labels for this region. This method achieves promising performance on one dataset. \citeauthor{cuevas2011circle} \cite{cuevas2011circle} propose discrete DE to search for three edge points to detect circles from images. The edge points are found by using the Canny detectors and DE is used to find three points, which can represent a circle in the image. This method can locate objects in complicated or noisy images.
 \citeauthor{ugolotti2013particle} \cite{ugolotti2013particle} optimise the model parameters in two model-based object detection systems, in one case using PSO and DE in the other. The two models solve two distinct tasks, i.e., human body pose estimation and hippocampus localisation in histological images. These two methods are implemented in CUDA to improve computational efficiency. \citeauthor{mussi2009gpu} \cite{mussi2009gpu, mussi2010gpu} develop a real-time PSO-based method in CUDA that finds the locations of road signs based on the search of the parameters of the perspective transform of a set of key contour points and their shape and colour information. %\citeauthor{song2019improved} \cite{song2019improved} propose the HS-NMS method using a harmony search algorithm to find good combinations of bounding boxes generated by CNN-based object detection methods and non-maximum suppression to focus only on the most relevant boxes. This method finds solutions by optimising an integrated function that considers the number and the quality scores (i.e., precision and recall) of the bounding boxes. This method is combined with two CNNs, i.e., region-based CNNs and region-based fully convolutional networks for object detection. The results show that this method improves detection and {\color{blue}localisation} performance over the original method. 

Salient object detection is a special case of object detection that only detects the most important object in an image and ignores other irrelevant ones. \citeauthor{singh2014novel} \cite{singh2014novel} propose a PSO method to find the optimal weights associated to three features and obtain a saliency map for object detection. A fitness function evaluating the attention pixels and the background pixels is used to guide the PSO search. This method outperforms existing methods according to different performance measures. \citeauthor{iqbal2016learning} \cite{iqbal2016learning} propose a learning classifier system-based method to evolve weights to linearly combine nine different feature maps for salient object detection. \citeauthor{afzali2017supervised} \cite{afzali2017supervised} propose a PSO method to find the optimal weights for combining nine feature maps with the goal of minimising the error rate. This method achieves better performance than five other methods on various images. \citeauthor{moghaddam2021automatic} \cite{moghaddam2021automatic} propose a GP-based method to construct high-level features from low-level saliency features for salient object detection. The feature sets are divided into four subsets and a GP method is used to construct features from each of them, which are combined to obtain the final saliency map. Unlike the PSO-based methods, GP-based methods perform non-linear combinations of the saliency feature maps. %More works related with salient object detection can be found in \cite{afzali2020evolutionary}.

\subsection{Automatically Evolving Detectors} 
EC methods have also been applied to automatically evolving detectors of objects in images. The main methods to evolve detectors are based on GP. Early works can be found in \cite{howard1999target, howard2006pragmatic, bhanu2004object, zhang2003domain}. % \citeauthor{howard1999target} \cite{howard1999target} propose a two-stage GP method to achieve target detection in SAR imagery. In the first stage, the GP method is used for detectors that detect target pixels from the chosen ocean pixels and the GP method in the second stage evolves programs to detect the clear targets and the false detections from the first stage. The results show that GP can automatically evolve a number of small and effective detectors. \citeauthor{howard2006pragmatic} \cite{howard2006pragmatic} develop a GP method with two types of terminals, i.e., pixel statistics from lines and circles and texture features extracted by discrete Fourier transform (DFT), respectively, to automatically evolve object and target detectors. The results show that the DFT terminals are more effective than pixel statistics in GP. \citeauthor{bhanu2004object} \cite{bhanu2004object} propose a GP method with 16 primitive images as terminals and 17 image processing operators as functions to evolve solutions for object detection. A soft tree size limit is proposed in the crossover operation to control the growth of tree size. This method achieves promising results on real synthetic aperture radar images. In \cite{lin2005object}, the performance of the GP method is further improved by developing a minimum description length-based fitness function that integrates the tree size and the detection rate. This method achieves better performance than standard GP for object detection on real synthetic aperture radar images.
%\citeauthor{zhang2003domain} \cite{zhang2003domain} propose a domain-independent multi-class object detection method based on GP. The work proposes a domain-independent feature extraction method using GP to evolve object detectors operating as classifiers based on these features. The evolved detectors scan the whole image and identify regions of interest as objects.
\citeauthor{zhang2007improving} \cite{zhang2007improving} proposes a new GP-based object detection method by investigating three feature sets, developing a new fitness function and a two-phase learning method. In the first phase, GP-based detectors are evolved and evaluated using a small subset of the training samples with the objective of maximising classification accuracy. In the second phase, the evolved detectors are further refined by using all training samples with the objective of maximising detection accuracy. The results show that the two-stage GP method is more stable and effective than a single-stage GP method for object detection. \citeauthor{liddle2010multi} \cite{liddle2010multi} propose a multi-objective GP method simultaneously maximising true positive rate and true negative rate based on a two-stage learning scheme that detects shapes or coins in images. This method can evolve an effective diverse set of object detectors. \citeauthor{li2021re} \cite{li2021re} propose a GP method to evolve a similarity measure for person re-identification. This method achieves better performance than a method based on the Euclidean distance. \citeauthor{olague2022automated} \cite{olague2022automated} apply the BP method \cite{olague2017brain} to automatically evolve a set of operators to generate saliency map and combine them to achieve salient object detection. This method achieves better performance than other EC-based methods and commonly used methods on two datasets.

To sum up, compared with image segmentation, image feature analysis, and image classification, there are fewer works on EC for object detection. Compared with these tasks, object detection is typically more difficult since it requires conducting both object localisation and object classification. However, EC methods have shown great potential in both optimising existing object detection systems and automatically evolving object detectors. But the potential of EC-based methods for object detection has not been comprehensively investigated. Future possible research directions may be investigating EC-based methods for evolving NNs, new EC methods with powerful representations to evolve detectors, and fast EC methods for object detection. Furthermore, the applications of EC-based methods to well-known object detection tasks such as COCO or real-world problems are desirable. 

\section{Other Tasks}
EC methods have shown great potential in other image-related tasks, including interest point detection, image registration, remote-sensing image classification, and object tracking. In this section, we will provide a brief review of works that have used EC for such applications.

\subsection{Interest Point Detection}
Interest points detection implies detecting interesting points such as corners, blobs, and edges that convey interesting or salient visual information. GP has been applied to automatically evolve interest point detectors in images. Most works are from Gustavo Olague's group. \citeauthor{trujillo2006synthesis} \cite{trujillo2006synthesis} develop a GP method containing arithmetic functions and image operators, such as histogram normalisation and Gaussian derivatives, to construct detectors from images. A new fitness function that measures detectors' repeatability, global separability, and information content is developed to guide the GP search. This method evolves two detectors, which are simpler but achieve better results than one hand-crafted solution. \citeauthor{trujillo2008automated} \cite{trujillo2008automated} further investigate the GP-based interest point detection method by analysing the effectiveness and robustness of 15 constructed detectors. The task of interest point detection often involves multiple potentially conflicting objectives, i.e., stability, point dispersion and information content. In \cite{olague2011evolutionary}, a multi-objective GP method optimising the objectives of stability and point dispersion is investigated for interest point detection. Two new detectors are evolved and analysed to show their effectiveness. \citeauthor{olague2012interest} \cite{olague2012interest} develop a multi-objective GP method to simultaneously optimise three objectives, which measure stability, point dispersion and information content. The results show that the non-dominated solutions found by GP outperform manually designed detectors.

\subsection{Image Registration}

Image registration is an important preprocessing task in many applications such as medical imaging. The task aims to align two or more images by finding an optimal transformation of the images in the geometric space. %The task is challenging since images can be sampled from different sensors, times, viewpoints, and environments, meaning the variations of images are high. 
EC methods have been developed for image registration, in particular finding the optimal transformation, where most of the works are from Cord\'on, Damas, and their collaborators and can be found in the survey papers \cite{santamaria2011comparative, valsecchi2013evolutionary}. %\citeauthor{bermejo2018coral} \cite{bermejo2018coral} present the coral reef optimization algorithm with substrate layers (CRO-SL) method for both feature-based and intensity-based medical image registration. 
In \cite{santamaria2012self}, a self-adaptive evolutionary optimization (SaEvO) algorithm is introduced to simultaneously search for the image registration solutions and the control parameters. %\citeauthor{bermejo2015comparative} \cite{bermejo2015comparative} propose several variants of BFA and provide comprehensive comparisons between these variants and existing EC methods for image registration. 
\citeauthor{gomez20183d} \cite{gomez20183d} present several methods including a DE method to search for optimal parameters in a 3D-2D silhouette-based image registration system. In the above methods, EC techniques have been used to find the optimal parameters of transformation models in image registration, achieving promising results. However, since EC techniques have flexible representations and powerful search ability and the EC-based applications to image registration have not been extensively investigated in recent years, there is still research potential in this direction.

\subsection{Remote Sensing Image Classification}
Remote sensing or hyperspectral image classification typically aims to classify each pixel in the images into different groups, not the entire image. Many EC-based methods have also been developed within this context, performing tasks like finding optimal cluster centres for the clustering-based methods, feature subset or band selection, and parameter optimisation. 
%\citeauthor{zhao2021spectral} \cite{zhao2021spectral} propose a GA-based band selection method for hyperspectral image classification. The main peculiarity of this GA method is the use of an unsupervised fitness function based on the within-class and between-class distances. This method achieves better accuracy than the standard GA method and seven other methods. 
\citeauthor{mai2021hybrid} \cite{mai2021hybrid} propose a PSO-based method to optimise the cluster centres and parameters in the interval type-2 semi-supervised possibilistic fuzzy C-means clustering method for satellite image classification. This method is tested on the tasks of landcover classification and landcover change detection. The results show that this method achieves better accuracy than most of the methods to which it is compared. %\citeauthor{maulik2010automatic} \cite{maulik2010automatic} propose a DE-based method to find the best number of clusters and cluster central for fuzzy c-means clustering for remote sensing image classification. 
\citeauthor{wang2017remote} \cite{wang2017remote} propose an ACO method with a binary encoding to simultaneously find optimal SVM parameters and select a subset of the 76 original features including Gabor wavelet, grey-level co-occurrence matrix(GLCM), histogram statistics. %The ACO method uses a binary encoding to represent the solutions, where each SVM parameter is encoded by a 10-bit binary string and the selection of a single feature is encoded by one bit. 
This method achieves better performance than other EC-based methods. %\citeauthor{shi2020evolutionary} \cite{shi2020evolutionary} propose an evolutionary multitask feature selection method to select a subset of spectral features for hyperspectral image classification. This method finds optimal feature subsets for building different individual classifiers and constructs an ensemble for final classification. 
\citeauthor{liu2020multiobjective} \cite{liu2020multiobjective} propose four different EMO-based remote sensing image classification methods to prune ResNet by removing filters, simultaneously minimising classifier complexity and maximising classification accuracy. \citeauthor{rs14051275} \cite{rs14051275} apply GP to evolve classifiers based on spectral, shape and texture features for cropland field classification on very high resolution images. This method achieves better performance than five traditional classifiers on two datasets. %\citeauthor{he2020remote} \cite{he2020remote} propose an adaptive ACO method to automatically evolve classification rules for remote sensing image classification. The ACO method achieves higher accuracy than PCA, RBF neural networks and other popular methods. 

\subsection{Object Tracking}
Object tracking aims to track objects in a series of images/frames by locating the objects' positions. Typically, a target object is given and the task is to find its position in multiple consecutive frames. \citeauthor{kang2018hybrid} \cite{kang2018hybrid} propose a hybrid gravitational search algorithm (HGSA) to search for the locations of objects by maximising a cosine-similarity function that evaluates the similarity between the target objects and potential objects in the image using features extracted by CNNs. This method improves the accuracy of a state-of-art SI-based object tracker. \citeauthor{song2013understanding} \cite{song2013understanding} propose a GP method to evolve detectors for tracking moving targets on a stationary background. This GP method uses pixels in a sliding window as terminals and a number of operators/functions to evolve the detectors/classifiers. The proposed method shows promise under different scenarios such as noise conditions. %\citeauthor{nyirarugira2013adaptive} \cite{nyirarugira2013adaptive} propose an adaptive DE method to search for the locations of gesturing objects for object tracking. Colour histograms are extracted from the detected objects as features and a similarity function is used to determine whether the objects are detected or not. This method also considers the motion blur information, improving the tracking efficiency and accuracy. 
\citeauthor{yan2021lighttrack} \cite{yan2021lighttrack} propose an EC-based one-shot object tracking NAS method to search for optimal trackers based on a pre-trained supernet. This method is more effective and efficient than the original methods without NAS search.

%\subsection{Others}
%Except for the above tasks, EC methods have been applied to a large number of image processing and analysis tasks, such as image outlier detection, image registration, image enhancement, image denoising, etc. In this section, we will show typical examples of EC methods for solving these different tasks. 

%\citeauthor{chen2018evolutionary} \cite{chen2018evolutionary} propose an image outlier detection method based on autoencoders and EMO. The image outlier detection task is to remove outliers that have different semantics from a set of images. This task is important to construct a good training dataset. In this method, EMO is used as an optimisation method to find the best penalty parameters of sparse group lasso in autoencoders by minimising the reconstruction loss and a regularisation term. The knee point from the final Pareto front is used to build AdaBoost-autoencoder to perform outlier detection. \citeauthor{li2020cascaded} \cite{li2020cascaded} propose an EA method as a data argumentation technique to generate 3D human pose image data using the crossover and mutation operators. 

%As an image processing task, image enhancement changes the pixel values of images in order to improve or generate better-quality images for specific applications. \citeauthor{bhandari2017new} \cite{bhandari2017new} propose a beta DE method to search for optimal values of gray levels to improve the quality and brightness of satellite images. 

\section{Applications, Challenges, and Trends}
This section summarises the application areas, the challenges and future trends of EC-based image analysis methods. Due to the page limit, information regarding the software/tools/packages and datasets related with EC for CV and image analysis have been included in the supplementary materials.

\subsection{Applications}

Existing EC-based image analysis methods have been applied to a variety of areas, including

\begin{enumerate}
%\item Texture analysis, including texture description, texture classification, and texture segmentation. 

\item Facial image analysis, e.g., face recognition \cite{bi2021multi}, facial expression classification \cite{ghazouani2021genetic}, ear detection \cite{ganesh2014entropy}, and eye detection \cite{abdel2014efficient}.

\item Biomedical image analysis, e.g., skin cancer classification \cite{ain2017genetic}, breast cancer classification \cite{ryan2015image}, prostate cancer classification \cite{thangavel2014soft}, brain tumour classification \cite{hemanth2019modified}, lung nodule classification \cite{luckehe2018evolutionary}, leukocyte segmentation \cite{saraswat2013leukocyte}, retinal vessel segmentation \cite{wei2021genetic}, cell image segmentation \cite{jiang2002evolutionary}, and ultrasound image segmentation \cite{rogai2016metaheuristics}.

\item Remote sensing image analysis, e.g., land cover classification \cite{mai2021hybrid}, scene classification \cite{liu2020multiobjective}, cropland field classification \cite{rs14051275}, hyperspectral image classification \cite{ghamisi2014feature},
satellite image segmentation \cite{suresh2017multilevel}, and seasonal change detection of riparian zones \cite{makkeasorn2009seasonal}.

\item Image analysis related to humans, e.g., action recognition \cite{liu2015learning}, body pose estimation \cite{ugolotti2013particle}, motion and human detection \cite{song2013understanding}.

\item Agricultural image analysis, e.g., flower classification \cite{bi2021genetic} and leaf disease detection \cite{singh2019sunflower}.

\item Others, e.g.,  motor imagery classification \cite{kirar2020combination}, art classification \cite{bi2020effective}, digit recognition \cite{bi2021divide}, kinship verification \cite{alirezazadeh2015genetic}, fish classification \cite{lin2005evolutionary}, and vehicle detection \cite{howard2006pragmatic}.
\end{enumerate} 

%In these applications, EC methods have been used for feature selection, feature extraction, parameter optimisation, classifier learning, or CNN architecture evolving, etc.

%\subsection{Texture Image Analysis}
%
%\subsection{Facial Image Analysis}
%
%\subsection{Medical Image Analysis}
%
%\subsection{Remote Sensing/Satellite Image Analysis}
%
%\subsection{Others}

\subsection{Challenges}
Despite many successful EC applications to image analysis, there are still significant issues and challenges as follows.

\subsubsection{Scalability}
Scalability is a common issue in most EC-based methods for image analysis. In recent years, big data have become a trend in image analysis, where the number of images in the datasets is very large (see the datasets summarised in the supplementary materials). Some well-known image/object classification datasets include the CIFAR10, CIFAR100 and ImageNet datasets. For instance,  CIFAR10/CIFAR100 comprises 60,000 32$\times$32 colour images and ImageNet more than 14 million images. There are also well-known large-scale image datasets for segmentation and object detection, such as the Berkeley Segmentation Dataset  and Microsoft COCO. Dealing with such large datasets using EC-based methods is often time-consuming and resource-intensive. Powerful computing devices such as GPUs may shorten execution time, but the computational cost is still very high if a large number of fitness evaluations are conducted, as happens when the search space is very large and characterised by many local optima.

\subsubsection{Representation} A careful design of representations is essential to the success of EC-based image analysis. Several different representations are used in EC methods, including string-based, vector-based, tree-based, and graph-based, allowing EC methods to solve a variety of image-analysis tasks. Even within the same specific task, EC methods may use multiple representations. For example, strings, vectors and graphs can be used as a representation for image feature selection \cite{xue2015survey}. Since representations are typically task-dependent, their design is challenging. In addition, they are highly related to the search space and the search mechanism, which are also key factors to a successful search. Due to their flexibility, however, there is still great potential for developing powerful EC representations for specific CV and image analysis tasks. 

\subsubsection{Search Mechanisms}
The underlying search mechanisms are at the heart of EC methods. Search mechanism means evolutionary operators, heuristics, and/or strategies that can make the search towards good directions and alternately (close to) optimal/good solutions in a (more) effective and efficient way. A good search mechanism can better balance exploration and exploitation and find globally optimal solutions. Image analysis tasks are typically very difficult, requiring powerful search mechanisms. Some works improve the search mechanisms of EC-based image analysis methods \cite{tarkhaneh2019adaptive, hemanth2019modified, zhang2022evolutionary}. EC methods are very flexible to be cooperated with other strategies, heuristics and operators to further improve their search. In addition, multi-objective search mechanisms are much less explored than single-objective ones in EC-based image analysis methods. Future work can consider these two directions to further explore the potential of EC methods in CV and image analysis.

\subsubsection{Interpretability}
Interpretability is very important in many image analysis tasks, such as biomedical image analysis. The most popular CV and image analysis methods in the last decade have been based on DNNs \cite{liu2021survey}, including DNNs optimised or designed by EC methods. A well-recognised drawback of DNN-based methods is their ``black-box" input/output mapping, often resulting in poor interpretability of their results. It is hard to explain how DNNs solve an image analysis task, e.g., why certain image features are selected/extracted and why the models are effective. Compared with DNNs, traditional methods relying on domain knowledge are typically more transparent and interpretable. However, traditional methods are affected by poorer performance, domain knowledge requirements, and poor flexibility. A trade-off can be found by making the use of domain knowledge more flexible and automatic by using EC methods. A typical example is to use GP to automatically evolve models with image filters and operators \cite{bi2021gpimage, shao2013feature}, which reduces the reliance on domain experts and improves the performance. More importantly, the models using image operators are more interpretable than DNNs. However, explainable AI using EC \cite{bacardit2022intersection} is a topic on which research is just starting blossoming. 

\subsubsection{Computational Cost}
Computational cost is an essential factor for EC-based image-analysis applications. Compared with random or exhaustive search methods, EC methods are by far less computationally expensive. In some cases, their powerful search ability makes them even faster than traditional CV methods. For example, in thresholding-based and clustering-based image segmentation methods, EC methods have shown their advantages in reducing computational cost while achieving results that are competitive with traditional methods \cite{abd2019many, zhao2020semisupervised}. However, in many supervised learning tasks, such as image classification, EC methods may be computationally expensive. Some attempts to improve the computational efficiency of EC-based image analysis have used surrogates to predict fitness values \cite{sun2019surrogate} or a small subset of a larger training set \cite{bi2021instance} for fitness evaluation. However, this topic has seldom been explored and therefore requires further work on improving computational efficiency of EC methods without sacrificing performance. 

\subsubsection{Recognition by the Main CV Community and Publishing Papers in Major CV and AI Conferences}
Despite the above technical challenges, other important challenges include the awareness of the contributions of EC-based methods by the CV community and publishing EC-based works in major CV and AI conferences, such as CVPR, ECCV, ICCV, IJCAI, and AAAI. For the EC-based contributions to be fully recognised by the major CV community, it is necessary to publish at these major conferences. However, this is very challenging due to the low acceptance rates. It is worth mentioning that some EC-based methods have been published in these conferences, such as \cite{yan2021lighttrack, li2020cascaded, lu2020nsganetv2}. But the total number of EC-related publications is very small, e.g., only about 0.24\% of CVPR publications out of 12724 publications include some EC-based mechanism (from Scopus). To popularise ECV, it is urgent to address this challenge. One chance might be offered by finding a relevant application/problem that EC can effectively solve but other AI techniques cannot, e.g., evolutionary NAS and EMO for image analysis.

\subsection{Trends}
In this section, several popular research trends of EC for CV and image analysis are summarised. 

\subsubsection{Evolutionary Deep Learning (EDL)}
The field of EDL studies the combinations of EC and deep learning \cite{al2019survey}, which is typically accomplished in two forms, i.e., using EC methods to optimise deep learning models and using EC methods to automatically evolve deep models from scratch. Most of the existing EDL methods belong to the first group, e.g., evolving deep CNN architectures for image segmentation and image classification \cite{liu2021survey}. A few GP-based methods belong to the second group, comprising studies in which GP has been used to automatically evolve deep models for image classification \cite{bi2021genetic, evans2018evolutionary}. The topic is of great interest and open to significant future developments due to the limitations of EC methods, the limitations of deep learning methods, and the wide variety of image analysis tasks that have been explored only very partially up to now. 

% \subsubsection{Integration of Domain Knowledge in EC Methods} Domain knowledge is often needed in many EC-based image analysis approaches. For example, domain knowledge is needed to extract effective features from images, before applying EC for feature selection \cite{alirezazadeh2015genetic} \cite{hemanth2019modified}. When domain knowledge is available, it is worth using it to help EC-based search for high-quality solutions by reducing the search space \cite{sengupta2019improved} \cite{fu2016genetic} \cite{bi2018genetic}, designing an effective fitness function \cite{fu2019bayesian} \cite{muangkote2017rr} \cite{alirezazadeh2015genetic}, improving the search mechanisms \cite{zhao2016multilevel} \cite{bhandari2020novel}, etc. The integration of domain knowledge into EC methods can improve the performance but often results in poor flexibility of the approaches, which are then hard to transfer to new tasks. Therefore, both the performance and the flexibility should be considered when integrating domain knowledge into EC. Since there are many real-world scenarios where domain knowledge is not available, it is advisable not to develop methods that rely too much on deep domain knowledge but general EC-based image analysis approaches, which can be easily applied for different tasks or specialised by adding domain knowledge, when available for specific tasks. %This is usually achieved by choosing specific fitness functions for general EC-based approaches.

\subsubsection{Computationally Efficient ECV Methods}
Improving the computational efficiency becomes important and popular in EC-based image analysis methods because image data are often large, high-dimensional, and complex. Computationally efficient EC methods have been studied from different perspectives, e.g., building surrogate models for fitness approximation \cite{bi2021instance, sun2019surrogate} and implementing EC methods on GPUs \cite{ugolotti2013particle}. Most of the existing computationally efficient EC methods are not originally designed for image analysis and they have a difficulty being transferred directly to image analysis because of differences in representations, search mechanisms, and tasks. Future work can be aimed at making EC-based image analysis methods, and particularly large-scale image analysis, more efficient and practical. 

\subsubsection{EMO for CV and Image Analysis}
Many CV and image analysis tasks typically have multiple (partially) conflicting objectives, for example, the model performance and the model complexity, or the number of features and the classification accuracy. EMO is very suitable for solving these tasks by providing non-dominated solutions with good tradeoffs between different objectives. Existing works show the potential of EMO in CV and image analysis \cite{liu2021multiobjective, zhao2020semisupervised, liu2020multiobjective}, but the potential can be further explored by designing different objective functions, developing powerful and task-dependent EMO search mechanisms, non-dominated solution selection or combination strategies, etc. 

\subsubsection{EC with Transfer Learning for CV and Image Analysis}
Transfer learning aims to extract knowledge from some (source) tasks and reuse it to improve the learning of other (target) tasks. CV and image analysis problems are often related or share similar characteristics. Transfer learning can be explored in EC-based image analysis methods to improve performance. Transfer learning can occur in a multitask learning manner \cite{bi2021learning} or a sequential transfer learning manner \cite{iqbal2017cross}. The knowledge can be transferred between the same type of tasks or different types of tasks \cite{bi2022multitask}. It is worth investigating how to effectively use the knowledge learned from one task in EC to improve its performance on another task.

\subsubsection{ECV Using Small-Scale Training Data} Few-shot learning or data-efficient learning has been recently a hot topic \cite{wang2020generalizing}, which focuses on learning from a few training instances. Current popular DNNs often rely on large-scale training data to achieve satisfactory performance in image-related tasks. However, dealing with small training data is also necessary, since labelling huge data can be extremely expensive and learning from them is very time-consuming. In this scenario, ECV methods may have a potential to achieve DNN-competitive results and overcome the limitations of DNNs \cite{bi2021dual}. 

\subsubsection{EC Methods for Interpretable CV and Image Analysis} 
Improving interpretability of the models/solutions is important for model reuse and acceptance, and poor interpretability is a drawback of DNN-based methods. EC methods can be used to provide more interpretable solutions by evolving high-quality but small-size models. However, this direction has not been fully explored in EC-based CV and image analysis methods. Future work can further explore how to reduce solution complexity, how to effectively evolve small-size solutions without affecting performance, etc.

\section{Conclusions}
This paper provides a comprehensive survey on recent works of EC techniques for CV and image analysis problems covering all essential EC techniques (e.g., GAs, GP, PSO, ACO, DE, and EMO) and major CV and image analysis tasks, including edge detection, image segmentation, image feature analysis, image classification, object detection, interest point detection, image registration, remote sensing image classification, and object tracking. This survey shows how these EC methods are applied to different image-related tasks by using/developing different encoding, search strategies and objective functions. More importantly, the limitations, challenges and issues of the reviewed works are discussed to provide more insights into this field. %Future directions and trends were presented to

Since 1970s, the field of ECV and image analysis has made great progress. By reviewing the related works in recent years, it is evident that EC-based methods have been applied to many different types of image-related tasks and showed great potential. %In particular, during the last decade, under the shade of deep learning methods, %particularly CNNs, which are dominant approaches in CV and image analysis tasks, %EC-based CV and image analysis methods are likely difficult to beat those methods.
Considering the recent history of the field, one can see that deep learning methods, especially CNNs, are dominant approaches in CV and image analysis, whose performance is hardly approached by the corresponding EC-based CV and image analysis methods. However, even those popular CNN-based methods have limitations, such as having poor interpretability, requiring rich expertise in the NN domain, and requiring expensive GPU computing resources. To address these limitations, it is necessary to develop effective, efficient and interpretable AI approaches to CV and image analysis. EC-based methods can be a good choice to address those limitations. For instance, EC-based NAS approaches reduce the requirement of expertise in the NN domain. In spite of the good performances frequently exhibited by EC-based methods, they still have to face several challenges and limitations in terms of scalability, representations, search mechanisms, computational cost, and interpretability. In parallel, the outbreak of new AI methods have fostered the emergence of new trends in EC-based image analysis, i.e., EDL, computationally efficient ECV methods, multi-objective approaches, EC with transfer learning, ECV using small-scale training data, and EC for interpretable CV and image analysis. It is desirable to develop new and powerful EC-based methods that provide effective but interpretable solutions to CV and image analysis tasks without expensive computing resources. This wide picture shows how EC-based CV and image analysis still has big potential for new research and offers great opportunities for new groundbreaking discoveries.

%In spite of the good performances frequently exhibited by EC-based methods, they still have to face several challenges and limitations in terms of domain knowledge integration, scalability, computational cost, representations, search mechanisms, and interpretability. In parallel, the outbreak of new AI methods have fostered the emergence of new trends in EC-based image analysis, i.e., evolutionary deep learning, computationally efficient EC methods, multi-objective approaches, and EC with transfer learning. This wide picture shows how EC-based image analysis still has big potential for new research and offers great opportunities for new groundbreaking discoveries. 

\bibliographystyle{IEEEtran}
\bibliography{IEEEabrv,survey.bib}
\end{document}

\title{Supplementary Materials of Paper ``A Survey on Evolutionary Computation for Computer Vision and Image Analysis: Past, Present, and Future Trends"}

\author{Ying Bi,~\IEEEmembership{Member,~IEEE}, Bing Xue,~\IEEEmembership{Senior Member,~IEEE}, Pablo Mesejo, Stefano Cagnoni,~\IEEEmembership{Senior Member,~IEEE}, Mengjie Zhang,~\IEEEmembership{Fellow,~IEEE}, 
\thanks{The author Ying Bi is with School of Electrical and Information Engineering, Zhengzhou University, Zhengzhou, China, and also with School of Engineering and Computer Science, Victoria University of Wellington, Wellington, New Zealand (e-mail: ying.bi@ecs.vuw.ac.nz).}% <-this % stops a space
\thanks{The authors Bing Xue and Mengjie Zhang are with School of Engineering and Computer Science, Victoria University of Wellington, Wellington, New Zealand (e-mail: bing.xue@ecs.vuw.ac.nz; mengjie.zhang@ecs.vuw.ac.nz).}% <-this % stops a space
\thanks{The author Pablo Mesejo is with Department of Computer Science and Artificial Intelligence (DECSAI), University of Granada, and also with the Andalusian Research Institute in Artificial Intelligence (DaSCI) and Panacea Cooperative Research S.Coop. Granada, Spain (e-mail: pmesejo@ugr.es).}
\thanks{The author Stefano Cagnoni is with Department of Engineering and Architecture, University of Parma, Italy (e-mail: stefano.cagnoni@unipr.it).}
\thanks{Color versions of one or more of the figures in this paper are available
online at http://ieeexplore.ieee.org.}}
\markboth{IEEE transactions on evolutionary computation,~Vol.~XX, No.~X, Month~year}
{Bi \MakeLowercase{\textit{et al.}}: paper title}
\maketitle

\section{Software and Datasets}
In this section, the commonly used software/packages/tools related to evolutionary computation (EC) and computer vision (CV) are summarised. The most commonly used datasets of CV and image analysis tasks are summarised as well. We hope this will be helpful for researchers, undergraduate and postgraduate students to conduct their research in this field.

\subsection{Packages/Software Related to Evolutionary Computation}
Most commonly used packages/software/tools in C++, Java, Python, Matlab, and R to implement EC algorithms including single-objective algorithms and multi-objective algorithms are summarised in Table \ref{table:EAsofteware}. Their links are also provided for easy use in the future. 

\begin{table*}[htbp]
	%	\vspace{-4mm}
	\footnotesize
	\setlength{\tabcolsep}{0.4em} % for the horizontal padding
	\caption{Summary of commonly used EC software/tools}
	\vspace{-4mm}
	\begin{center}
		\begin{tabular}{|p{0.15\linewidth}|p{0.06\linewidth}|p{0.41\linewidth}|p{0.34\linewidth}|}
			\hline 
			Name&Language&Description&Link \\ \hline
			Open BEAGLE \cite{gagne2006genericity}&C++ &A generic EC software &\url{https://github.com/chgagne/beagle}\\
			ECF&C++&Evolutionary computation framework&\url{http://ecf.zemris.fer.hr/}\\
			EALib&C++&An evolutionary algorithms library&\url{https://github.com/dknoester/ealib}\\
			SharpNEAT&C++&A complete implementation of neuroevolution of augmenting topologies &\url{https://sharpneat.sourceforge.io/}\\
			Darwin Neuroevolution Framework&C++&A framework intended to make Neuroevolution experiments easy, quick and fun&\url{https://github.com/tlemo/darwin}\\
			EO \cite{keijzer2001evolving}&C++&Evolving objects: an evolutionary computation framework&\url{http://eodev.sourceforge.net/}\\
			pagmo2 \cite{Biscani2020}&C++&A C++ platform to perform parallel computations of optimisation tasks (global and local) via the asynchronous generalized island model&\url{https://github.com/esa/pagmo2}\\
			% &&&\\
			\hline

			ECJ \cite{luke2006ecj}&Java&A Java-based evolutionary computation research system&\url{http://cs.gmu.edu/~eclab/projects/ecj/}\\
			JCLEC&Java&Java class library for evolutionary computation&\url{http://jclec.sourceforge.net/}\\
			Watchmaker Framework&Java&An extensible, high-performance, object-oriented framework for implementing platform-independent evolutionary/genetic algorithms in Java&\url{https://watchmaker.uncommons.org/}\\
			MOEA Framework&Java&A free and open source framework for multiobjective optimization&\url{http://moeaframework.org/}\\
			Jenetics&Java&A genetic algorithm, evolutionary algorithm, grammatical evolution, genetic programming, and multi-objective optimization library&\url{https://github.com/jenetics/jenetics}\\
			JGAP&Java&A genetic algorithms and genetic programming package&\url{https://sourceforge.net/projects/jgap/}\\
			jMetal&Java&A framework for multi-objective optimization with metaheuristics&\url{https://jmetal.github.io/jMetal/}\\ 
			\hline

			DEAP \cite{fortin2012deap}&Python&Distributed evolutionary algorithms in Python&\url{http://code.google.com/p/deap/}\\
			Pyvolution&Python&Evolutionary algorithms framework&\url{https://pypi.org/project/Pyvolution/1.0/}\\
			LEAP \cite{coletti2020library}&Python&Library for evolutionary algorithms in Python&\url{https://leap-gmu.readthedocs.io/en/latest/#}\\
			pySTEP&Python&Python strongly typed genetic programming&\url{http://pystep.sourceforge.net/}\\
			pyeasyga&Python&A simple and easy-to-use implementation of a genetic algorithm library in Python&\url{https://pypi.org/project/pyeasyga/}\\
			Geatpy2&Python&Evolutionary algorithm toolbox and framework with high performance for Python&\url{https://github.com/geatpy-dev/geatpy}\\
			Evol&Python&A python grammar for evolutionary algorithms and heuristics&\url{https://github.com/godatadriven/evol}\\
			pymoo \cite{pymoo}&Python& Multi-objective optimization in Python&\url{https://github.com/anyoptimization/pymoo}\\
			\hline

			% &&&\\
			% &&&\\
			GEATbx&Matlab&Genetic and evolutionary algorithm toolbox&\url{http://www.geatbx.com/docu/index.html}\\
			YPEA&Matlab&A toolbox for evolutionary algorithms in MATLAB&\url{https://yarpiz.com/}\\
			PlatEMO \cite{PlatEMO}&Matlab&Evolutionary multi-objective optimization platform&\url{https://github.com/BIMK/PlatEMO}\\
			ev-MOGA&Matlab&Multiobjective evolutionary algorithm&\url{https://www.mathworks.com/matlabcentral/fileexchange/31080-ev-moga-multiobjective-evolutionary-algorithm}\\
			% &&&\\
			\hline

			ECR&R&Evolutionary computation in R, inspired by DEAP&\url{https://cran.r-project.org/web/packages/ecr/vignettes/introduction.html}\\
			% &&&\\
			% &&&\\
			% &&&\\			
			\hline		
		\end{tabular}	
		\label{table:EAsofteware}
	\end{center}
	\vspace{-6mm}
\end{table*}

\subsection{Packages/Software Related to Computer Vision}

CV and image analysis is also a big field with many packages and tools. The popular ones are summarised in Table \ref{table:CVsofteware}. It is worth mentioning that OpenCV is the most popular one among all these tools/packages and it supports different programming languages such as C++, Java, Python, and Matlab. In addition, several deep learning packages such as CAFFE, YOLO, Keras, Tensorflow, PyTorchCV, and torchVision have been widely used to deal with different CV and image analysis tasks, which are also  summarised in Table \ref{table:CVsofteware}.

\begin{table*}[htbp]
		% \vspace{-4mm}
	\footnotesize
	\setlength{\tabcolsep}{0.4em} % for the horizontal padding
	\caption{Summary of commonly used CV, image processing and analysis software/package}
	\vspace{-4mm}
	\begin{center}
		\begin{tabular}{|p{0.1\linewidth}|p{0.11\linewidth}|p{0.44\linewidth}|p{0.3\linewidth}|}
			\hline 
			Name&Language&Description&Link \\ \hline
			OpenCV&C++, Python, Java, MATLAB&An open source library for image processing and computer vision&\url{https://opencv.org/}\\
			VIGRA&C++&Vision with generic algorithms&\url{https://ukoethe.github.io/vigra/}\\
			CAFFE& C++ &A fast open framework for deep learning&\\
			YOLO&C, Python&A latest and cutting edge real-time object detection system&\url{https://pjreddie.com/darknet/yolo/}\\

			\hline

			BoofCV&Java&An open source library written from scratch for real-time computer vision&\url{http://boofcv.org/index.php?title=Main_Page}\\
			JavaCV&Java&Java interface to OpenCV, FFmpeg, and more&\url{https://github.com/bytedeco/javacv}\\
			ImageJ&Java&A public domain software for processing and analysing scientific images&\url{https://imagej.net/}\\
			% &Java&&\url{}\\
			\hline
			Imutils&Python&A computer vision package&\url{https://pypi.org/project/imutils/}\\
			IPSDK&C++ and Python&An image processing library&\url{https://www.reactivip.com/image-processing/}\\
			torchVision&Python&Computer vision models on PyTorch&\url{https://pytorch.org/vision/stable/index.html}\\
			PyTorchCV&Python&A collection of image classification, segmentation, detection, and pose estimation models&\url{https://pypi.org/project/pytorchcv/}\\
			Scikit-Image &Python&A collection of algorithms for image processing&\url{https://scikit-image.org/}\\
			SimpleCV&Python&An open source framework for building computer vision applications&\url{http://simplecv.org/}\\
			Keras&Python&A Python-based open-source software library that acts as an interface for the machine learning platform TensorFlow&\url{https://keras.io/}\\
			Tensorflow&Python&TensorFlow is an end-to-end open source platform for machine learning&\url{https://www.tensorflow.org/}\\
			DeepFace&Python&A lightweight face recognition and facial attribute analysis (age, gender, emotion and race) framework&\url{https://github.com/serengil/deepface}\\
			OpenVINO&Python&An open-source toolkit for optimizing and deploying AI inference&\url{https://docs.openvino.ai/latest/index.html}\\

			\hline
			MatImage&Matlab&A Matlab library for analysis and processing of digital images&\url{https://github.com/mattools/matImage}\\
			Image Processing Toolbox&Matlab&Perform image processing, visualisation, and analysis&\url{https://www.mathworks.com/products/image.html}\\

			% &Java&&\url{}\\

			\hline		
		\end{tabular}	
		\label{table:CVsofteware}
	\end{center}
	\vspace{-6mm}
\end{table*}

\subsection{Commonly Used Computer Vision Datasets}

EC techniques have been applied for dealing with different CV and image analysis tasks. For better practice, Tables \ref{table:datasets} and \ref{table:datasets22} summarise the popular CV and image analysis datasets, including tasks of edge detection, image segmentation, image classification, and object detection. These datasets have also been used to test the performance of the EC-based methods in the literature. The references and links of these datasets are provided and more information can be found via checking them accordingly and easily. %Note that other CV and image analysis datasets are not included since they are not very commonly used in EC-related literature. 

\begin{table*}[htbp]
	%	\vspace{-4mm}
	\footnotesize
	\setlength{\tabcolsep}{0.4em} % for the horizontal padding
	\caption{Summary of commonly used datasets solved by using EC methods. These datasets are typically related to \textbf{one type of task}}
	%\vspace{-4mm}
	\begin{center}
		\begin{tabular}{|p{0.18\linewidth}|p{0.14\linewidth}|p{0.37\linewidth}|p{0.3\linewidth}|}
			\hline 
			Dataset&Task&Description&Link\\ \hline	
			% \multicolumn{4}{|c|}{\textbf{Datasets for a Single Task}}\\ \hline
			DRIVE: Digital Retinal Images for Vessel Extraction&Vessel segmentation&20 training images and 20 testing images&  \url{https://drive.grand-challenge.org}\\
			Weizmann horses \cite{borenstein2004learning}&Image segmentation&328 colour images of horses &  \url{https://www.msri.org/people/members/eranb/}\\	
			SBD (Semantic Boundaries Dataset) \cite{hariharan2011semantic}&Image segmentation&11318 images and masks of object instance boundaries&  \url{http://home.bharathh.info/pubs/codes/SBD/download.html}\\

			CIFAR-100 \cite{krizhevsky2009learning}&Object Classification&60000 32$\times$32 colour images in 100 classes. 50000 images for training and 10000 images for testing&  \url{https://www.cs.toronto.edu/~kriz/cifar.html}\\		

			CIFAR-10 \cite{krizhevsky2009learning}&Object Classification&60000 32$\times$32 colour images in 10 classes. 50000 images for training and 10000 images for testing&  \url{https://www.cs.toronto.edu/~kriz/cifar.html}\\	
			MNIST \cite{lecun1998gradient}&Digit Classification&60000 28$\times$28 gray-scale images for training and 10000 28$\times$28 gray-scale images for testing in 10 classes&  \url{http://yann.lecun.com/exdb/mnist/}\\
			MNIST variants &Digit Classification&Several variants of MNIST with adding more variations such as background and noise&  \url{https://sites.google.com/a/lisa.iro.umontreal.ca/public_static_twiki/variations-on-the-mnist-digits}\\	
			Fashion-MNIST \cite{xiao2017fashion}&Object Classification&70000 28$\times$28 gray-scale images of 10 fashion products. 60000 images for training and 10000 images for testing& \url{https://github.com/zalandoresearch/fashion-mnist}\\
			STL-10 \cite{coates2011analysis}&Image Classification&13000 labelled 96$\times$96 images from 10 object classes & \url{https://cs.stanford.edu/~acoates/stl10/}\\		
			Caltech-101 \cite{fei2004learning}&Object Classification&A set of images in 101 classes. Each class has about 50 images& \url{https://data.caltech.edu/records/20086}\\
			SVHN \cite{netzer2011reading}&Digit classification&600000 32$\times$32 RGB images of digits in 10 classes&\url{http://ufldl.stanford.edu/housenumbers/}\\
			The EMNIST Dataset&Image classification&A set of 28$\times$28 images with handwritten character digits & \url{https://www.nist.gov/itl/products-and-services/emnist-dataset}\\
			CINIC-10 \cite{darlow2018cinic}&Image classification&An extension of CIFAR-10 with 270000 images in 10 classes&  \url{https://github.com/BayesWatch/cinic-10}\\
			Food-101 \cite{bossard14}&Image classification&A dataset of 101 food categories with 101000 images&  \url{https://data.vision.ee.ethz.ch/cvl/datasets_extra/food-101/}\\
			FGVC-Aircraft \cite{maji13finegrained}&Object classification&A benchmark dataset contains 10200 images of aircraft with 100 images for each of 102 different aircraft model variants &\url{https://www.robots.ox.ac.uk/~vgg/data/fgvc-aircraft/}\\
			Oxford-IIIT Pets \cite{parkhi2012cats}&Object classification&A 37 category pet dataset with roughly 200 images for each class&  \url{https://www.robots.ox.ac.uk/~vgg/data/pets/}\\
			Oxford 102 Flower \cite{nilsback2008automated}&Object classification&A 102 category dataset consisting of 102 flower categories&  \url{https://www.robots.ox.ac.uk/~vgg/data/flowers/102/}\\
			CUB-200-2011 \cite{wah2011caltech}&Object classification&11788 bird images of 200 subcategories& \url{https://deepai.org/dataset/cub-200-2011}\\	
			COIL20 \cite{nene1996columbia}&Object classification&20 different objects with 72 images&  \url{https://www.cs.columbia.edu/CAVE/software/softlib/coil-20.php}\\
			COIL100 \cite{nene1996columbia}&Object classification&100 different objects with 72 images&  \url{https://www1.cs.columbia.edu/CAVE/software/softlib/coil-100.php}\\
			Brodatz Textures&Texture classification&112 texture images&  \url{https://www.ux.uis.no/~tranden/brodatz.html}\\	
			% Promise12 &&The task is to segment the prostate in transversal T2-weighted MR images.&  \url{https://promise12.grand-challenge.org}\\
			KTH-TIPS and KTH-TIPS2 &Texture classification&A collection of texture images with variations in scale, pose and illumination&  \url{https://www.csc.kth.se/cvap/databases/kth-tips/}\\
			Describable Textures Dataset (DTD) \cite{cimpoi14describing}&Texture classification&A texture database consisting of 5640 images in 47 classes&  \url{https://www.robots.ox.ac.uk/~vgg/data/dtd/}\\		
			Kylberg texture \cite{kylberg2011kylberg}&Texture classification&28 texture classes and 160 576$\times$576 unique texture patches per class& \url{https://kylberg.org/kylberg-texture-dataset-v-1-0/}\\	
			UCF101 \cite{soomro2012ucf101}&Action recognition&Consists of 13320 video clips in 101 categories&  \url{https://www.crcv.ucf.edu/data/UCF101.php}\\
			Scene data \cite{lazebnik2006beyond}&Scene classification&Scene images in 15 classes&  \url{https://figshare.com/articles/dataset/15-Scene_Image_Dataset/7007177}\\		
			Places \cite{zhou2017places}&Scene classification&2.5 millions of images in 205 scene categories&  \url{http://places.csail.mit.edu}\\	
			Multi-PIE \cite{gross2010multi}&Face recognition&Contains more than 750000 images of 337 people &\url{https://www.cs.cmu.edu/afs/cs/project/PIE/MultiPie/Multi-Pie/Home.html}\\		
			Color FERET \cite{phillips2000feret}&Face recognition&11338 512$\times$768 colour images with 13 different poses from 994 subjects& \url{https://www.nist.gov/itl/products-and-services/color-feret-database}\\
			UMIST&Face recognition&564 images of 20 individuals&\url{http://eprints.lincoln.ac.uk/id/eprint/16081/}\\		
			ORL \cite{samaria1994parameterisation}&Face recognition&400 92$\times$112 images from 40 subjects&  \url{https://cam-orl.co.uk/facedatabase.html}\\	
			Yale Face Database &Face recognition&165 gray-scale images of 15 individuals & \url{http://vision.ucsd.edu/content/yale-face-database}\\	
			Extended Yale B \cite{georghiades2001few}&Face recognition&2414 frontal-face images with size 192$\times$168 from 38 subjects &  \url{http://vision.ucsd.edu/~leekc/ExtYaleDatabase/ExtYaleB.html}\\
			Faces95 &Face recognition&1440 180$\times$200 images from 72 individuals&  \url{https://cmp.felk.cvut.cz/~spacelib/faces/faces95.html}\\
			JAFFE (Japanese Female Facial Expression) \cite{lyons1998coding}&Facial expression classification& 213 images of different facial expressions from 10 different Japanese female subjects& \url{https://zenodo.org/record/3451524#.YrOfMS-6H5g}\\
			Cohn–Kanade (CK) dataset&Facial expression classification&8795 face images of six universal facial expressions from 97 different subjects&  \url{http://www.jeffcohn.net/Resources/}\\		
			MSRA Salient Object Database \cite{HouPami19Dss} \cite{HouCvpr2017Dss}&Salient object detection&A database containing over 10000 images&  \url{https://mmcheng.net/msra10k/}\\	
			INRIA Person \cite{dalal2005histograms}&Object Detection&614 person detections for training and 288 for testing&\url{https://dbcollection.readthedocs.io/en/latest/datasets/inria_ped.html}\\
			\hline		
		\end{tabular}	
		\label{table:datasets}
	\end{center}
	%\vspace{-6mm}
\end{table*}

\begin{table*}[htbp]
		\vspace{-4mm}
	\footnotesize
	\setlength{\tabcolsep}{0.4em} % for the horizontal padding
	\caption{Summary of commonly used datasets solved by using EC methods. Each of these datasets include \textbf{multiple types of tasks}}
	\vspace{-4mm}
	\begin{center}
		\begin{tabular}{|p{0.18\linewidth}|p{0.14\linewidth}|p{0.37\linewidth}|p{0.3\linewidth}|}
			\hline 
			Dataset&Task&Description&Link\\ \hline	
			% \multicolumn{4}{|c|}{\textbf{Datasets for Multiple Tasks}}\\ \hline
			Berkeley Segmentation Dataset and Benchmark \cite{MartinFTM01}&Edge detection, image segmentation&12000 hand-labeled segmentations of 1000 Corel dataset images from 30 human subjects& \url{https://www2.eecs.berkeley.edu/Research/Projects/CS/vision/bsds/} \\		
			Berkeley Segmentation Dataset and Benchmarks 500 (BSDS500)&Edge detection, image segmentation&300 training images and 200 testing images &\url{https://www2.eecs.berkeley.edu/Research/Projects/CS/vision/grouping/resources.html}\\	
			York Urban Line Segment Database Information&Edge detection, image segmentation&102 640$\times$480 images (45 indoor, 57 outdoor) of urban scenes from the campus of York University and downtown Toronto, Canada& \url{https://www.elderlab.yorku.ca/resources/york-urban-line-segment-database-information/}\\	
			NYUv2 (NYU-Depth V2) \cite{silberman2012indoor}&Edge detection, segmentation&1449 densely labeled pairs of aligned RGB and depth images and 407024 unlabeled frames&  \url{https://cs.nyu.edu/~silberman/datasets/nyu_depth_v2.html}\\	
			Outex texture classification dataset&Segmentation, classification&Consists of 17 texture classification test suites& \url{https://computervisiononline.com/dataset/1105138685}\\
			&&&  \url{}\\	
			Dermofit Image Library&Segmentation, classification&A collection of high quality skin lesion images &  \url{https://licensing.edinburgh-innovations.ed.ac.uk/product/dermofit-image-library}\\
			PH$^2$ Database&Segmentation, classification&A total of 200 dermoscopic images of melanocytic lesions, including 80 common nevi, 80 atypical nevi, and 40 melanomas&  \url{https://www.fc.up.pt/addi/ph2%20database.html}\\	
			UC Merced Land Use Dataset&Segmentation, classification&A 21 class land use image dataset. The size of images is 256$\times$256 & \url{http://weegee.vision.ucmerced.edu/datasets/landuse.html}\\	
			Cell Tracking Challenge&Image segmentation, object tracking&The task includes segmenting and tracking moving cells in time-lapse video sequences&\url{http://celltrackingchallenge.net}\\
			RIM-ONE \cite{inproceedings2015Fumero}&Segmentation, classification&159 stereo eye fundus images with a resolution of 2144$\times$1424& \url{https://www.idiap.ch/software/bob/docs/bob/bob.db.rimoner3/stable/index.html}\\	
			ImageNet \cite{deng2009imagenet}&Object classification, detection&Includes 14197122 annotated images& \url{https://www.image-net.org}\\		
			COCO (Microsoft Common Objects in Context) \cite{lin2014microsoft}&Object detection, segmentation, keypoint detection&Includes 328000 images with 1.5 million object instances& \url{https://cocodataset.org/#home}\\		
			&&&  \url{}\\
			Pascal VOC datasets \cite{everingham2015pascal}&Segmentation, classification, detection&A challenge of datasets for different tasks& \url{http://host.robots.ox.ac.uk/pascal/VOC/index.html}\\		
			\hline		
		\end{tabular}	
		\label{table:datasets22}
	\end{center}
	\vspace{-6mm}
\end{table*}

% Insight Segmentation and Registration Toolkit (ITK)

% S. Klein, M. Staring, K. Murphy, M.A. Viergever, J.P.W. Pluim Elastix: a toolbox for intensity-based medical image registration IEEE Trans. Med. Imag., 29 (1) (2010), pp. 196-205, 10.1109/TMI.2009.2035616

\bibliographystyle{IEEEtran}
\bibliography{survey.bib}